Article

# Generalizing AI-driven Assessment of Immunohistochemistry across Immunostains and Cancer Types: A Universal Immunohistochemistry Analyzer

Biagio Brattoli[1,+], Mohammad Mostafavi[1,+], Taebum Lee[1,+], Wonkyung Jung[1], Jeongun Ryu[1], Seonwook Park[1], Jongchan Park[1], Sergio Pereira[1], Seunghwan Shin[1], Sangjoon Choi[2], Hyojin Kim[3], Donggeun Yoo[1], Siraj M. Ali[1], Kyunghyun Paeng[1], Chan-Young Ock[1], Soo Ick Cho[1,*], and Seokhwi Kim[4,5*]

[+]These authors contributed equally to this work.
[*]Corresponding authors

[1]Lunit, Seoul, Republic of Korea

[2]Department of Pathology and Translational Genomics, Samsung Medical Center, Sungkyunkwan University School of Medicine, Seoul, Republic of Korea

[3]Department of Pathology, Seoul National University Bundang Hospital, Seongnam, Republic of Korea

[4]Department of Pathology, Ajou University School of Medicine, Suwon, Republic of Korea

[5]Department of Biomedical Sciences, Ajou University Graduate School of Medicine, Suwon, Republic of Korea

Correspondence to:

Seokhwi Kim, M.D., Ph.D.,

Department of Pathology, Ajou University School of Medicine.

Department of Biomedical Sciences, Ajou University Graduate School of Medicine.

164 Worldcup-ro, Yeongtong-gu, Suwon, 16499, Republic of Korea.

Tel: +82-31-219-6460; Fax: 82-31-219-5934; E-mail: seokhwikim@ajou.ac.kr

ORCID ID: 0000-0001-7646-5064

Soo Ick Cho, M.D., Ph.D.,

Lunit.

374 Gangnam-daero, Gangnam-gu, Seoul, 06241, Republic of Korea.

Tel: +82-2-2138-0827; Fax: 82-2-6919-2702; E-mail: sooickcho@lunit.io

ORCID ID: 0000-0003-3414-9869

**ABSTRACT**

Despite advancements in methodologies, immunohistochemistry (IHC) remains the most utilized ancillary test for histopathologic and companion diagnostics in targeted therapies. However, objective IHC assessment poses challenges. Artificial intelligence (AI) has emerged as a potential solution, yet its development requires extensive training for each cancer and IHC type, limiting versatility. We developed a Universal IHC (UIHC) analyzer, an AI model for interpreting IHC images regardless of tumor or IHC types, using training datasets from various cancers stained for PD-L1 and/or HER2. This multi-cohort trained model outperforms conventional single-cohort models in interpreting unseen IHCs (Kappa score 0.578 vs. up to 0.509) and consistently shows superior performance across different positive staining cutoff values. Qualitative analysis reveals that UIHC effectively clusters patches based on expression levels. The UIHC model also quantitatively assesses c-MET expression with MET mutations, representing a significant advancement in AI application in the era of personalized medicine and accumulating novel biomarkers.

**Introduction**

Immunohistochemistry (IHC) is an antibody-based methodology that can reveal the expression and distribution of proteins in formalin-fixed paraffin-embedded (FFPE) tissues and is well established as a decision support tool for oncology diagnosis[1,2]. IHC results are now increasingly used to guide decision making for systemic therapy for disseminated malignancy such as for the monoclonal antibody pembrolizumab in non-small cell lung cancer (NSCLC) as based on Programmed Death-Ligand 1 (PD-L1) expression[3,4]. Moreover, multiple emerging classes of therapies based on monoclonal antibodies (antibody-drug conjugates [ADC], bi-specific antibodies) directly target proteins on the tumor cell surface[5,6]. The efficacy of these cell surface-targeting therapeutics is consistently linked with the expression of the targeted protein. Therefore, quantifying IHC assessments of these targets may facilitate the development of predictive biomarkers that are valuable in clinical practice[7].

Recently, artificial intelligence (AI) models have been developed to quantify IHC images by tissue segmentation, cell delineation, and quantification of all relevant cells in a whole slide image (WSI)[8,9]. However, the development of these AI models is heavily constrained by their reliance on single training cohorts that typically contain at least several hundred or often more WSI cases of a cancer type and immunostains matched to the desired indication. Moreover, these training sets are manually labeled on a cellular/subcellular basis by pathologists with each slide taking several hours for annotation depending on complexity[10,11].

Importantly, there is an additional limitation of 'domain-shift', where current deep-learning models for IHC cannot recognize elements - either immunostain for cancer type - that are not present in the training set. This limitation indicates for each immunostain-cancer type combination, an IHC training set must be created and annotated with accompanying significant time and resource cost, which is particularly relevant when evaluating new antibodies for development[12,13]. Both the requirement for expert annotated training sets specific to each desired permutation of immunostain and cancer type and the domain shift problem intertwine to create an imperative for a universally applicable AI model that is proficient in interpreting IHC results without antecedent manually annotated matching training sets[14].

Here, we developed a Universal IHC (UIHC) analyzer, which can assess IHC images, irrespective of the specific immunostain or cancer type. Eight models trained on WSI patches from three cancer types, immunostained for PD-L1 or human epidermal growth factor receptor 2 (HER2), were defined by exposure to varying single or multiple cohorts for training. Models trained on single cohorts served as the benchmark, whereas models trained with multiple cohorts were an innovation[10,15,16]. All models were evaluated using a diverse test set including eight 'novel' IHC stained cohorts covering twenty additional cancer types, along with two 'training' IHC (PD-L1 and HER2) stained cohorts to identify the best model for further development.

**Results**

**Patch-level tumor cell detection and IHC-positivity classification**

We trained both single-cohort-derived models (SC-models) with one dataset and multiple-cohort-derived models (MC-models) with multiple datasets based on NSCLC, urothelial carcinoma, and breast cancer datasets stained with PD-L1 22C3 and breast cancer datasets stained with HER2 (Fig. 1). Fig. 2a shows the combination of different datasets to develop the eight AI models utilized in this study. SC-models exhibit favorable performance within test sets matched for the immunostain and cancer type used for training as evidenced by the cell detection (negatively stained Tumor Cell [TC-] or positively stained Tumor Cell [TC+]) performance (median F1-score [min, max]) of P-L (PD-L1 22C3 of lung) on the PD-L1 22C3 Lung test set (0.693 [0.686, 0.705], Fig. 2b), P-Bl (PD-L1 22C3 of bladder) on the PD-L1 22C3 Bladder test set (0.725 [0.719, 0.731], Fig. 2c), P-Br (PD-L1 22C3 of breast) on the PD-L1 22C3 Breast test set (0.599 [0.590, 0.607], Fig. 2d), and H-Br (HER2 of breast) on the HER2 test set (0.759 [0.753, 0.766], Fig. 2e). Notably, MC-models with broader exposure beyond the matched training set (P-LBlBr [PD-L1 22C3 for lung, bladder, and breast], PH-Br [PD-L1 22C3 and HER2 for breast], PH-LBr [PD-L1 22C3 and HER2 for lung and breast], PH-LBlBr [PD-L1 22C3 and HER2 for lung, bladder, and breast]) performed as well as or better than the best performing SC-model for each test set matched to a training set, regardless of immunostain or cancer type. (Fig. 2b-e).

For test sets containing novel elements that were not seen in training, MC-models significantly outperformed SC-models. Notably, for the test set with an experienced immunostain but unseen cancer types, such as PD-L1 22C3 Pan-cancer set in Fig. 2f, MC-models trained with more cancer types (P-LBlBr) and/or an additional stain (PH-LBr and PH-LBlBr) outperformed the SC-models P-L (P-LBlBr, 0.722 [0.716, 0.730], $p<0.001$; PH-LBr, 0.745 [0.735, 0.753], $p<0.001$; PH-LBlBr, 0.743 [0.735, 0.752], $p<0.001$), which were the best performing SC-models.

In the other novel cohorts with unseen immunostains such as PD-L1 SP142, Claudin 18.2, Delta-like 3 (DLL3), fibroblast growth factor receptor 2 (FGFR2), human epidermal growth factor receptor 3 (HER3), mesenchymal-epithelial transition factor (MET), MUC16, and trophoblast cell-surface antigen 2 (TROP2), MC-models generally performed better than SC-models Fig. 2g-n). Most representatively identified in MET Pan-cancer, all the MC-models outperformed the single best performing SC-model H-Br (P-LBlBr, 0.795 [0.773, 0.810], p<0.001; PH-Br, 0.762 [0.725, 0.776], *p<0.001*; PH-LBr, 0.783 [0.767, 0.799], *p<0.001*; PH-LBlBr, 0.792 [0.776, 0.815], *p<0.001*) (Fig. 2l). This tendency for MC-models to outperform SC-models was also observed when the data was categorized by cancer type (lung, breast, bladder, pan-ovary, esophagus, colorectum, and stomach) rather than IHC type (Supplementary Fig. 1).

**WSI-level IHC quantification of MC- and SC-models**

The performances of the AI models at the WSI level were subsequently assessed using the test sets outlined in Supplementary Table 1. The ground truth images were categorized and annotated based on the tumor proportion score (TPS), and the models' performance was evaluated by accurately assigning the WSIs to the corresponding ground truth group (TPS <1%; 1-49%; ≥50%). Among the eight models, the PH-LBlBr model was the top performer for this set of test WSI cohorts, achieving a Cohen's kappa score of 0.578 and an accuracy of 0.751 (Fig. 3a, Supplementary Fig. 2a). The best SC-model overall was H-Br, but it still had significantly lower performance, with a Cohen's kappa score of 0.509 and an accuracy of 0.703. In assessing performance on the PD-L1 22C3 Lung WSI test set, PH-LBlBr was the only model to outperform the SC-model P-L, with a Cohen's kappa score of 0.652 compared to 0.638 for P-L, and an accuracy of 0.793 compared to 0.785 for P-L (Fig. 3b, Supplementary Fig. 2b). For the PD-L1 22C3 Pan-cancer WSI test set and the PD-L1 SP142 Lung WSI test set, P-LBlBr also outperformed all SC-models, including P-L (Fig. 3c-d, Supplementary Fig. 2c-d). Notably, in the multi-stain pan-cancer test set, the PH-LBlBr model consistently outperformed all SC-models (P-L, P-Bl, P-Br, H-Br) and MC-models with less diversity in training cohorts (PH-Br, P-LBlBr, and PH-LBr), achieving a Cohen's kappa score of 0.610 and an accuracy of 0.757 (Fig. 3e, Supplementary Fig. 2e). Confusion matrices indicated that the PH-LBlBr model performed evenly across different TPS levels, with the highest number of concordance cases and mispredictions predominantly falling into adjacent categories (e.g., fewer mispredictions of TPS<1% as TPS≥50%) (Fig. 4a-b). Due to its consistently high performance across test sets, PH-LBlBr is designated as the UIHC model.

**Performance analysis of UIHC on novel immunostains for different cutoffs**

For certain immunostains commonly utilized in clinical practice, such as MET, TROP2, and MUC6, the absence of consensus scoring systems poses a challenge. To evaluate the false and true positive rates for these immunostains in our analysis, we initially established a cutoff at 1% to maintain a standardized ground truth (GT)-TPS, similar to the approach used for PD-L1 staining, while varying the AI model-predicted TPS cutoff. In this binary classification framework, the area under the receiver operating characteristics (AUROC) curve demonstrates that the selected UIHC model (92.1%) outperforms SC-Models (Fig. 5a). Additionally, we compared our AI models across a range of cutoffs from 1% to a second value within the range of [2%, 75%], illustrating a three-way classification accuracy of 78.7% (Fig. 5b). In both analyses, the UIHC model consistently exhibits superior performance, irrespective of the specific cutoff applied for novel stain types.

**Histopathologic validation of inference examples of UIHC model**

Representative discrepancy cases between the UIHC model and SC-models were subjected to WSI-level histopathological validation by pathologists (T.L., W.J., S.C., and S.K) to assess the accuracy of the models in detecting IHC-positive cells. In a case involving MET-stained NSCLC, the SC model P-L incorrectly classified the majority of tumor cells as negative (TPS 36%) (Fig. 6a). Conversely, the UIHC model accurately identified tumor cells based on positivity (TPS 61%), yielding results similar to the average TPS assessment of 75% by pathologists. In another instance concerning FGFR2-stained gastric cancer (Fig. 6b), the P-L model encountered difficulties, often failing to recognize numerous tumor cells and distinguish between FGFR2-positive and negative cells. In contrast, the UIHC model demonstrated an ability to discern IHC positivity even amidst this intricate staining pattern.

**Interpreting the representations of UMAP learned by the UIHC model**

To ensure the absence of inadvertent biases acquired during training, we evaluated the learned representations of the UIHC model using standard UMAP (uniform manifold and projection) for visualization. Two-dimensional internal representations of various AI models were presented in two formats: the 2D projection across training and novel cohorts (Fig. 7a), and a mosaic of image patches organized based on their respective projections (Fig. 7b).

In Fig. 7a, ground-truth TPS values were color-coded, transitioning from blue (0%) to brown (100%). For comparison, scatter plots were presented for three different sources: raw pixels as the baseline (Fig. 7a, left), features from a self-supervised learning (SSL) model trained with the same UIHC details but on larger, unannotated datasets (Fig. 7a, center), and the UIHC model (Fig. 7a, right). The pixel model (Fig. 7a, left) exhibited weak clustering signal, with high TPS patches clustered towards the bottom-right. In the SSL model (Fig. 7a, center), clustering based on TPS was not observed, but rather clustering based on cohort. The 2D projection depicted in Fig. 7a, right illustrated that the UIHC model effectively separated and clustered patches based on TPS expression level. Our visual inspections of UMAP mitigated the Clever Hans effect (skewing of results by external biases) commonly observed in machine learning[17]. This analysis effectively demonstrates that our approach facilitated the development of an AI-powered analyzer capable of generalizing to novel immunostains and cancer types, even in IHCs with cytoplasmic staining not included in the training data, indicating superior feature extraction through exposure to multiple cohorts.

Fig. 7b presents a mosaic of original image patches arranged according to their internal representation as observed in Fig. 7a. In contrast to raw pixels, the features of the UIHC model were centered around tumor cell detection and classification rather than visual attributes derived from varying source characteristics such as color contrast or brightness. Thus, the pixel representation prioritizes sorting by color, while the UIHC model remains unbiased by appearance, focusing instead on tumor type. Cohort similarity results further indicate that only the UIHC model exhibits reduced sensitivity to cohort-specific traits, indicating its lack of bias towards IHC type and emphasis on the primary task of detecting and classifying tumor cells (Fig. 7c).

**Performance analysis of UIHC on the real-world dataset**

To evaluate the UIHC model's applicability as a real-world assessment tool, we employed it to quantitatively assess the expression of c-MET, a novel immunostain for the model, in three cohorts of NSCLC cases known to harbor oncogenic driver alterations - MET exon 14 skipping mutations, MET amplifications, and epidermal growth factor receptor (EGFR) exon 20 insertions [18]. The UIHC model assigned higher tumor proportion scores (TPS) to the MET amplification group compared to the other groups (Table 1). The UIHC model yielded a MET TPS of 94.5±2.0 for the three MET amplification cases, 77.1±17.7 for the six exon 14 skipping mutation cases, and 75.7±23.2 for the seven EGFR exon 20 insertion EGFR cases.

**Discussion**

In this study, we demonstrated that UIHC trained with multiple cancer types and IHC, the MC-model, is not only superior in the domain used for training SC-model trained with a single cancer type and IHC in its domain but also exhibited the capability to analyze never-before-seen immunostains and cancer types.

Emerging therapeutic agents, meticulously designed to target surface proteins on tumor cells, have exerted a profound influence on the landscape of oncology care. These therapeutics can be broadly categorized into targeting tumor-associated antigens (TAA, such as TROP2) and targeting immune checkpoints (IC, such as PD-L1)[19-21]. Specific examples include trastuzumab deruxtecan, an ADC targeting HER2, and tarlatamab, a bispecific molecule targeting DLL3 and CD3[22-24].

IHC stands as an essential component in cancer diagnosis, and thus far, the pathologist's reading remains the gold standard for determining the expression level of a target protein[4,25-29]. Nonetheless, discrepancies between pathologists and poor reproducibility can hinder precise evaluation[10,15,16,30-33]. Efforts have been made to standardize IHC assays to maintain its role as a predictive biomarker, requiring evaluations as quantitative as possible. Recently introduced deep learning models have exhibited notable advantages over traditional computational methods, primarily due to their capacity to discern intricate patterns within IHC images where the latter requires the pathologists to understand the tissue structure and morphology directly per case before analysis[34,35]. These models can analyze PD-L1 and HER2 expression but require training on a large, manually annotated training cohort[10,11,15,16,36-38]. Moreover, such deep-learning models have domain shift issues that are effective within the cancer type and immunostain defined by the training cohort, but not for indications that contain cancer types and immunostains not within the training set[12,13,39].

In the present study, AI models underwent training using either a single cohort (SC) or multiple cohorts (MC). The MC-models, particularly those exposed to the most diverse range of cases, demonstrated superior performance compared to the SC-models. This was evident across test sets similar to the training cohorts, as well as test cohorts composed of previously unexposed (novel) immunostains and cancer types. The enhanced performance of MC-models in training cohorts can be attributed to the augmented training data. Compared to the H-Br model, the PH-Br model showed better performance on PD-L1 22C3 breast and HER2 breast, indicating the impact of increasing the volume of training data. However, the superiority of PH-Br over PH-LBr in PD-L1 22C3 bladder, which was trained with a larger cohort than PH-Br, suggests that the influence of expanding the training data volume is not straightforward. Irrespective of the volume of training data, training models using cohorts from various cancer types or immunostains together contributed to improve model performance. This phenomenon is exemplified in PD-L1 22C3 Pan-cancer, where PH-LBr, encompassing variations in both cancer type and immunostain, outperforms P-LBlBr, which only varies in cancer type, or PH-Br, which only varies in immunostain. The impact of variations in cancer type or immunostain within the training data is underscored by the superior performance of MC-models compared to SC-models, particularly evident for novel cohorts. Conversely, in the case of novel cohorts such as FGFR2 IHCs, where both membrane and cytoplasmic intensity can be observed, AI models trained solely on membranous staining IHCs (e.g., PD-L1 22C3 and HER2) may experience significant performance degradation[40,41]. Indeed, among the novel cohorts, both SC- and MC-models exhibited the poorest performance on FGFR. The UIHC model, however, demonstrated superior performance compared to other models, particularly in detecting cytoplasmic stained TC+, whereas most SC-models struggled to identify cytoplasmic stained TC+.

Recent AI-related research disciplines can be divided into the two main branches of model-centric and data-centric AI[42]. The model-centric AI focuses on designing and optimizing the best AI models with a fixed dataset, while data-centric AI systematically and algorithmically focuses on providing the best dataset for a fixed AI model. Our study underscores the promising efficacy of training the AI model with diverse IHC and cancer type data. Notably, this is clinically meaningful because it was done without additional data work, mostly annotation in a novel cohort, so it can be applied directly to new targets. Recent trends tend to call approaches with large training set from different domains 'foundational models', therefore, in this sense, our UIHC could be considered one[43-45]. However, we reserve this name for a multi-modal system that goes beyond histopathology and combines multiple medical disciplines[46].

To demonstrate the possible clinical utility of the current analyzer, we assessed c-MET expression in NSCLC to address the long-standing question of targeting c-MET. MET amplification is strongly believed to be correlated with increased expression of c-MET, however, so are exon 14 splicing mutations in c-MET (METex14m)[47,48]. Specifically, these mutations lead to the omission of exon 14 and the Cbl sites which are thought to be recognized by an E3 ubiquitin ligase, and thus thought to increase the amount of c-MET expressed by the tumor cell[49]. As theorized, c-MET amplifications lead to high expression of c-MET as seen in previous studies[47,50]. In contrast, tumors with METex14m had similar expression to exon 20 insertion NSCLC driven tumors. These unripe findings should be replicated in a larger cohort, but are very relevant to the development and clinical use of large molecular therapeutics targeting c-MET such as amivantanab[22].

There are some limitations in this work. The current scope of IHC expression detection is confined to tumor cells, but not other cell types, i.e. lymphocytes and macrophages. However, the UIHC model is able to learn to assess these other cell types if given the correct training sets as consistent with a data-centric approach. Furthermore, our IHC evaluation was limited to a binary categorization of positive or negative, but will encompass multi-level protein expression assessments such as the American Society of Clinical Oncology (ASCO) / College of American Pathologists (CAP) guidelines for HER2 in the future[51]. In addition, the model's performance demonstrated some variability across different staining techniques and cancer types within this study. This concern could potentially be addressed through the inclusion of additional IHC stain types within the model's training dataset, in other words exposing the model to more multiple cohorts in training.

In conclusion, we have successfully developed a UIHC model capable of autonomously analyzing novel stains across diverse cancer types. In contrast to prevailing literature and existing image analysis products that often focus on specialized cohorts, our model's versatility and agility significantly enhance its potential in expediting research related to new IHC antibodies[34]. This innovative approach not only facilitates a broad spectrum of novel biomarker investigations but also holds the potential to assist in the development of pioneering therapeutics.

**Methods**

**Dataset preparation for AI model development**

Histopathology dataset for annotation

The dataset used to develop the model consists of a total of 3,046 WSIs including lung (NSCLC), urothelial carcinoma, and breast cancer cases stained for PD-L1 22C3 pharmDx IHC (Agilent Technologies, Santa Clara, CA) and breast cancer WSIs stained for anti-HER2/neu (4B5) (Ventana Medical Systems, Tucson, AZ), as reported in previous studies (Fig. 1, Supplementary Table S2)[10,15,16]. All data for this study were obtained from commercially available sources from Cureline Inc. (Brisbane, CA, US), Aurora Diagnostics (Greensboro, NC, US), Neogenomics (Fort Myers, FL, US), Superbiochips (Seoul, Republic of Korea) or were available by the permission of Institutional Review Board (IRB) from Samsung Medical Center (IRB no. 2018-06-103), Seoul National University Bundang Hospital (IRB no. B-2101/660-30), and Ajou University Medical Center (IRB no. AJOUIRB-KS-2023-425). All slide images and clinical information were de-identified and pseudonymized.

The WSIs were divided into training, tuning (also called validation), and test sets. Since WSIs are too large for computation, a section of size $0.04mm^2$ (patch, i.e. tile) is extracted.

To evaluate and compare the models, we collected patch-level test sets from ten different stain types: PD-L1 22C3 (lung, bladder, breast, liver, prostate, colorectum, stomach, biliary tract, and pancreas), HER2 (breast), PD-L1 SP142 (lung), various immunostain types including Claudin 18.2, DLL3, FGFR2, HER3, MET, MUC16, and TROP2 (pan-cancer). The test sets of PD-L1 22C3 lung, bladder, and breast originated from the same cohort of training and tuning sets mentioned above (internal test set in Supplementary Table 2), which could be referred to as training domain. Patches of PD-L1 22C3 other than lung, bladder, and breast were from WSIs of colorectum (n = 19), liver (n = 20), stomach (n = 18), prostate (n = 18), pancreas (n = 19), and biliary tract (n = 20). For the novel domain test set, which is never shown to the AI model during training, we collect patches from novel cancer types and novel immunostain types. Additionally, patches of various immunostain types were from pan-cancer (more than 25 cancer types) tissue microarray (TMA) cores (Superbiochips, Seoul, Republic of Korea)[52-55]. Detailed information on antibodies for various immunostain types is provided in Supplementary Table 3. All slides were scanned by P1000 scanner (3DHistech, Budapest, Hungary) or Aperio AT2 scanner (Leica Biosystems Imaging, Buffalo Grove, IL, US). Within a WSI, up to three patches are selected and then resized to 1024x1024 pixels, at a normalized Microns-Per Pixel (MPP) of 0.19 µm. Such MPP normalization is required to unify the resolution of the patches since WSIs scanned from different scanners can have different MPP values. The patches are extracted manually to avoid uninteresting areas, such as the white background. No patches of the same WSI can be found in different sets, to prevent information leakage between the training and test sets.

Patch-level annotation for AI model development

We define two general cell classes for IHC by TC- or TC+ (Fig. 8a). In most of the IHC staining, except HER2, the expression was described as either positive or negative. Patches stained with HER2 are traditionally annotated with four levels of IHC quantification as follows; H0 (negative), H1+ (faint/barely perceptible and incomplete membrane staining), H2+ (weak to moderate complete membrane staining), and H3+ (complete, intense circumferential membrane staining)[56]. To unify the categories across stains, we remapped H0 to negatively stained Tumor Cell (TC−) and the remaining H1~H3 to positively stained Tumor Cell (TC+).

All annotations were performed by board-certified pathologists. The interpretation of tumor cell positivity by pathologists was determined by following the guidelines for PD-L1 or HER2[51,57]. The training set was composed of 574,620 TC+ and 1,415,033 TC−, while the tuning set contained 138,429 TC+ and 316,808 TC− (Supplementary Table 4, Fig. 8b). The tuning set was used to select the best checkpoint during the model training phase. The total TC+ and TC- annotated from the patches of the test set are described in Supplementary Table 5.

WSI-level test sets for AI model performance validation

Given that a single patch is a tiny fraction (<1%) of a WSI, performance of any model on the WSI-level can significantly deviate from patch-level assessment[58]. Therefore, a comprehensive comparison of our model performance on WSI was conducted with the key output of WSI-level TPS[59,60].

We collected four WSI-level test sets: PD-L1 22C3 lung (n = 479), PD-L1 22C3 pan-cancer (n = 135), PD-L1 SP142 lung (n = 178) and a novel, multi-stain test set (n = 140) as presented in Fig. 8c and Supplementary Table 1. The test set containing PD-L1 22C3 lung cancer was used in previous publications.(*10*) The PD-L1 22C3 Pan-cancer contains cancer types of biliary tract (n = 23), colorectum (n = 23), liver (n = 23), stomach (n = 23), prostate (n = 22), and pancreas (n = 21). The test set containing PD-L1 SP142 lung (n = 178) was derived from the same cohort of PD-L1 22C lung cancer. IHC in the multi-stain test set included MET, MUC16, HER3, TROP2, DLL3, FGFR2, Claudin 18.2, SP142, and E-Cadherin across ten cancer types. Except for PD-L1 22C3 lung cancer, they all corresponded to novel domains. Representative image samples from both training and novel groups are illustrated in

Supplementary Fig. 4 and 5.

The multi-stain test set contains following stains: Claudin18.2 (n = 18), DLL3 (n = 16), E-Cadherin (n = 10), FGFR2 (n = 18), HER3 (n = 15), MET (n = 25), MUC16 (n = 16), PD-L1 SP142 (n = 10), and TROP2 (n = 12) across ten cancer types (lung, breast, bladder, cervix, colorectum, esophagus, liver, lung, melanoma, stomach). Within the multi-stain test set, except for PD-L1 SP142 which is applied only on lung cancer, other staining antibodies (n = 130) are used for: stomach (n = 39), bladder (n = 28), breast (n = 23), lung (n = 19), cervix (n = 5), esophagus (n = 5), melanoma (n = 4), colorectum (n = 3), head and neck (n = 3), liver (n = 1).

TPS evaluation for all datasets was performed by three independent board-certified pathologists (S.C., H.K., and S.K. for PD-L1 22C3 lung, S.C., W.J., and S.K. for PD-L1 SP142 lung, and T.L., S.C., and S.K. for PD-L1 22C3 pan-cancer and multi-stain set).

**AI Model development process**

<u>Development of Universal IHC algorithm</u>

Our approach's inference pipeline consists of training dataset preparation, AI model development, and performance validation with diverse cohorts (Fig. 1). Specifically, after extracting patches and annotating cells from designated training cohorts, several AI models are trained with single-cohort (standard approach) or multiple-cohort data (innovation). Each model's parameters have been tuned using their domain-specific tuning (validation) set. Using combinations of the above cohorts, we produce eight models as described in Fig. 2a. While SC-models (H-Br, P-L, P-Bl, and P-Br) are trained on a single cohort[10,15,16]. MC-models such as P-LBlBr, PH-Br, PH-LBr, and PH-LBlBr are trained on multiple cohorts. Among these candidate models, we aim to identify the model that exhibits the highest degree of generalization for designation as a UIHC model.

Models are then tested on patches or WSIs exclusively held out from the training dataset. Our testing encompassed multiple cohorts, including 'training cohorts' and 'novel cohorts'. Most patches posed greater challenges as the staining proteins or cancer types were not part of the training data for any AI models presented in this study.

Label pre-processing

Inspired by the previous work that trains the cell detection model with point annotations, we define cell detection as a segmentation task[13,61]. At training time, we provide the cell labels as a segmentation map by drawing a disk centered on each cell point annotation. We use a fixed radius of ~ $1.3\mu m$, corresponding to 7 pixels at a resolution of 0.19 MPP. Finally, we assign the value of pixels within each disk based on the class of a cell, '1' for TC-, '2' for TC+. '0' is assigned for the remaining pixels.

Inference post-processing

Given that we treat cell detection as a segmentation task, a post-processing phase is needed to extract 2D coordinates and classes of predicted cells from the probability map output by the network. We apply *skim-age.feature.peak_local_max* on the model's output, which finds the locations of local maximums of the probability map to get the set of predicted cell points[62]. Lastly, we obtain each cell's class and probability value in the cell segmentation map through *argmax*. This probability is used as the confidence score.

Network architecture and training details

For all of our models, we use DeepLabV3+ as our base architecture with a ResNet34 encoder, which is a popular architecture specifically designed for the segmentation task[63,64]. During training, we augment the patches with a set of standard data augmentation methods for computer vision. In particular, we utilize center crop, horizontal and vertical flip, rotation, gaussian noise, color jittering, and gray scaling. Random values are sampled for each augmentation every time an image is loaded. Network parameters are initialized by Kaiming initialization[65]. The model is optimized using the Adam optimizer[66]. Dice loss is used to train the model[67]. The initial learning rate is set to $1e-4$, adapted using the cosine learning rate scheduler[68]. All the models have been trained for 150 epochs and evaluated at every 10 epochs to choose the best checkpoint on a hold-out tuning set. An epoch that shows the highest mF1 score on the tuning set is chosen as the best epoch and used for all evaluation purposes. All of the models are trained and evaluated with the same machine specifications as follows: 4 NVIDIA Tesla T4 GPUs each with 16GB of GPU memory and 216GB of RAM.

Inference details on whole slide images (WSIs)

For WSI inference we use the full WSI for tumor proportion score (TPS) calculation, excluding white background and in-house control tissue regions. The WSI is divided into 1024×1024 pixels of non-overlapping patches with an MPP of normalized 0.19 (following the training data), which are fed to our network, producing a prediction map with the same size as the input. All outputs are then combined to obtain a prediction map for the full WSI.

**AI Model evaluation**

Metrics for AI model performance

The performance evaluation of the AI model was analyzed at the patch-level and WSI-level. At the patch-level, performance was measured by F1 score, which compares the results of pathologists' annotation of each cell with the results of the AI model. At the slide-level, TPS by pathologists or UIHC was divided into categories based on a given cutoff threshold (1%/50% [3 classes]). Then performance was evaluated by comparing the TPS categories from pathologists to the AI model using Cohen's Kappa. The details of F1 score and TPS are described in the Supplementary methods.

Model interpretation by visualization of data distribution

To gain deeper insights into the learned patterns of the UIHC model, we delved into its inference process by extracting internal representations of the network for each image patch in the test set. We visualized these representations in 2D using UMAP, a widely used method for dimensionality reduction method for visualization[69]. We utilized a 2D projection where each point is a patch and the Euclidean distance between two points indicates the similarity within the network's internal representation. For this experiment, we developed two baselines to provide context for our UIHC:

1. Raw pixel representation, by simply downsizing the image from 1024×1024×3 to 32×32×3 and flattening the pixels, producing a 1×3072 vector. RGB-channel is kept since the color is important for IHC quantification. For the same reason this is a valid baseline, in fact, simply looking at the intensity of the brown color is a good indicator for TC-/+.

2. We developed a second network for comparing two deep learning models. Since patches with a manual annotation are a small fraction of all slides, we trained a ResNet34 using a state-of-the-art SSL method called Barlow Twins instead[70]. This allowed us to train the model on a large number of histopathology patches from different types of stains (PD-L1 22C3, and HER2) and cancer types without the need for any manual annotations.

3. The best UIHC model is used as the representative UIHC model for the qualitative analysis.

To extract the internal representation from the deep learning models (UIHC and SSL), each patch runs through the ResNet34 encoder producing a 16×16×512 tensor of shape Height×Width×Channels. The output tensor is averaged over spatial dimensions, thus producing a 1×512 vector. After producing a vector for each of the *N* patches in our test set, we obtain a matrix of *N* ×512 (*N* ×3072 for *Pixel*). Finally, we can project the 2858×512 matrix to *N* ×2 by using UMAP, a popular non-linear dimensionality reduction algorithm[69]. This 2-dimension matrix can be easily plotted as a scatter plot using *matplotlib* and *seaborn*. To calculate cohort similarities, we compute the Wilcoxon test between all cohort pairs, producing a similarity matrix of size [# cohorts × # cohorts] containing p-values. Then we average the upper-triangular matrix shown in the bar plot. In addition, the mosaic of image patch is drawn by discretizing the latent representations and replacing each point with the corresponding original patch image[71]. For each discretized point in space, the median patch is selected as the representative of that cluster.

Analysis of a genomically defined MET NSCLC dataset with the UIHC model

To further validate the performance of the UIHC model, we ran AI model inference on MET-stained NSCLC WSIs (n = 15) with gene mutation/amplification profiles. The cases were all diagnosed with NSCLC at Ajou University Medical Center and confirmed by next-generation sequencing to have either EGFR exon20ins, MET exon skipping, or MET amplification alterations.

**Reporting summary**

Further information on research design is available in the Nature Research Reporting Summary linked to this article.

**Data availability**

The processed data can be provided by the corresponding authors after formal requests and assurances of confidentiality are provided.

**Code availability**

Deep-learning-related code was implemented using *pytorch* version 1.12, Python version 3.9 and publicly available neural network architectures, like ResNet (open-source available online, e.g. https://github.com/pytorch/vision/blob/main/torchvision/models/resnet.py) and DeepLabV3 (open-source available online, e.g. https://github.com/VainF/DeepLabV3Plus-Pytorch). For UMAP we utilize the official, open-source implementation (available at

https://umap-learn.readthedocs.io/en/latest/clustering.html#using-umap-for-clustering). All plots were generated with publicly available libraries, matplotlib version 3.5.2 (available at https://github.com/matplotlib/matplotlib/tree/v3.5.2) and seaborn version 0.12.2 (available at https://seaborn.pydata.org/whatsnew/v0.12.2.html), using Google Colab (available at https://colab.google/).

**References**

1. Stack, E. C., Wang. C., Roman, K. A. & Hoyt, C. C. Multiplexed immunohistochemistry, imaging, and quantitation: a review, with an assessment of Tyramide signal amplification, multispectral imaging and multiplex analysis. *Methods* **70**, 46–58 (2014).

2. Cregger, M., Berger, A. J. & Rimm, D. L. Immunohistochemistry and quantitative analysis of protein expression. *Arch. Pathol. Lab. Med.* **130**, 1026–1030 (2006).

3. Slamon, D. J. et al. Use of Chemotherapy plus a Monoclonal Antibody against HER2 for Metastatic Breast Cancer That Overexpresses HER2. *N. Engl. J. Med.* **344**, 783–792 (2001).

4. Garon, E. B. et al. Pembrolizumab for the Treatment of Non–Small-Cell Lung Cancer. *N. Engl. J. Med.* **372**, 2018–2028 (2015).

5. Fuentes-Antrás, J., Genta S., Vijenthira, A. & Siu, L. L. Antibody–drug conjugates: In search of partners of choice. *Trends Cancer* (2023).

6. Qian, L. et al. The Dawn of a New Era: Targeting the "Undruggables" with Antibody-Based Therapeutics. *Chem. Rev.* **123**, 7782–7853 (2023).

7. Patel, S. P. & Kurzrock, R. PD-L1 expression as a predictive biomarker in cancer immunotherapy. *Mol. Cancer Ther.* **14**, 847–856 (2015).

8. Baxi, V., Edwards, R., Montalto, M. & Saha, S. Digital pathology and artificial intelligence in translational medicine and clinical practice. *Mod. Pathol.* **35**, 23–32 (2022).

9. Ibrahim, A. et al. Artificial intelligence in digital breast pathology: techniques and applications. *The Breast* **49**, 267–273 (2020).

10. Choi, S. et al. Artificial intelligence–powered programmed death ligand 1 analyser reduces interobserver variation in tumour proportion score for non–small cell lung cancer with
26

**Acknowledgments**

We appreciate Heeyeon Kay for the language editing. This work was supported by the National Research Foundation of Korea (NRF) grant funded by the Ministry of Science and ICT (MSIT) (2022R1C1C1007289), Republic of Korea, new faculty research fund of Ajou University School of Medicine, and Lunit.

**Author contributions**

B.B., M.M., S.P., S.S., D.Y., S.M.A., K.P, C-Y.O., S.I.C., and S.K. conceptualized the research. T.L., S.S., S.C., H.K., C-Y.O., S.I.C., and S.K. contributed to data acquisition. B.B. and M.M. conceived the experiments, B.B., M.M., J.R., and J.P. conducted the experiments. B.B., M.M., J.R., S.P., J.P., S.P, D.Y., and K.P. developed algorithms. B.B., M.M, T.L, J.R., S.P., S.P., S.C., H.K., S.I.C., and S.K. interpreted and validated the algorithms. B.B., M.M., S.I.C., S.M.A., and S.K. prepared an initial draft of the manuscript. All authors reviewed and approved the final version of the manuscript.

**Competing interests**

Biagio Brattoli, Mohammad Mostafavi, Taebum Lee, Jeongun Ryu, Seonwook Park, Jongchan Park, Sergio Pereira, Seunghwan Shin, Donggeun Yoo, Siraj M. Ali, Kyunghyun Paeng, Chan-Young Ock, and Soo Ick Cho are employees of Lunit and/or have stock/stock options in Lunit.

**Figure legend**

**Fig. 1. Overview of the universal immunohistochemistry (UIHC) artificial intelligence (AI) model development.** Single-cohort-derived models (SC-models) were trained using one dataset, while multiple-cohort-derived models (MC-models) were trained using multiple datasets, including lung, urothelial carcinoma, and breast cancer samples stained with Programmed Death-Ligand 1 (PD-L1) 22C3, as well as breast cancer samples stained with human epidermal growth factor receptor 2 (HER2). The AI models' performance was validated on both the training cohorts and novel cohorts that were not included in the training phase. These novel cohorts consisted of samples stained for human epidermal growth factor receptor 3 (HER3), MUC16, mesenchymal-epithelial transition factor (MET), trophoblast cell-surface antigen 2 (TROP2), and fibroblast growth factor receptor 2 (FGFR2).

**Fig. 2. Patch-level quantitative analysis of the artificial intelligence (AI) models. a** List of eight AI models trained on different cohort combinations. H-Br, HER2 of breast; P-L, PD-L1 22C3 of lung; P-Br, PD-L1 22C3 of breast; P-LBlBr, PD-L1 22C3 of lung, bladder, and breast; PH-B, PD-L1 22C3 and HER2 of breast; PH-LBr, PD-L1 22C3 and HER2 of lung and breast; PH-LBlBr, PD-L1 22C3 and HER2 of lung, bladder, and breast. The different stain combinations (e.g. PD-L1 or HER2 is utilized or not), are visualized by color. **b-e** Performance of the eight models in training cohorts, where the stain type may be utilized during training – **b** PD-L1 22C3 in lung cancer, **c** PD-L1 22C3 in bladder cancer, **d** PD-L1 22C3 in breast cancer, **e** HER2 in breast cancer, **f** PD-L1 22C3 in pan-cancer. **g-n** Performance of the eight models in novel cohorts - **g** PD-L1 SP142, **h** Claudin 18.2, **i** DLL3, **j** FGFR2, **k** HER3, **l** MET, **m** MUC16, **n** TROP2 - where none of the test immunostain types has ever been utilized during the training

phase by any of the models. PD-L1, programmed death-ligand 1; HER2, human epidermal growth factor receptor 2; DLL3, delta-like 3; FGFR2, fibroblast growth factor receptor 2; HER3, human epidermal growth factor receptor 3; MET, mesenchymal-epithelial transition factor; TROP, trophoblast cell-surface antigen; mF1, mean F1 score.

**Fig. 3. Whole slide image (WSI)-level quantitative analysis of the artificial intelligence (AI) models.** The quantitative analysis is based on comparing the tumor proportion score (TPS) score in different training settings. The reported Cohen's Kappa scores are computed using the pathologists' labeled category as ground truth. **a** Macro-averaged Cohen's Kappa scores of the eight AI models over all the stains. **b** Cohen's Kappa scores of the AI models in PD-L1 22C3 Lung dataset. **c** Cohen's Kappa scores of the AI models in PD-L1 22C3 Pan-cancer dataset. **d** Cohen's Kappa scores of the AI models in PD-L1 SP142 Lung dataset. **e** Cohen's Kappa scores of the AI models in multi-stain Pan-cancer dataset. The X-axis presents the summation of utilized stain types and the organ types of each cohort when training (e.g. PH-Br [PD-L1 22C3 and HER2 of breast] is 3 as it has 2 stains and 1 cancer type). H-Br, HER2 of breast; P-L, PD-L1 22C3 of lung; P-Br, PD-L1 22C3 of breast; P-LBlBr, PD-L1 22C3 of lung, bladder, and breast; PH-B, PD-L1 22C3 and HER2 of breast; PH-LBr, PD-L1 22C3 and HER2 of lung and breast; PH-LBlBr, PD-L1 22C3 and HER2 of lung, bladder, and breast.

**Fig. 4. Performance analysis of the artificial intelligence (AI) models on whole slide image (WSI) categorized by tumor proportion score (TPS). a** Confusion matrices of multiple-cohort-derived models (P-LBlBr [PD-L1 22C3 of lung, bladder, and breast], PH-Br [PD-L1 22C3 and HER2 of breast], PH-LBr [PD-L1 22C3 and HER2 of lung and breast], PH-LBlBr

[PD-L1 22C3 and HER2 of lung, bladder, and breast]). **b** Confusion matrices of single-cohort-derived models (H-Br [HER2 of breast], P-Br [PD-L1 22C3 of breast], P-L [PD-L1 22C3 of lung], and P-Bl [PD-L1 22C3 of bladder]). 1% and 50% were utilized as TPS cutoffs.

**Fig. 5. Performance analysis of the artificial intelligence (AI) models on novel immunostains with varying interpretation cutoffs. a** The receiver operating characteristic (ROC) curve by changing the cutoff over the predicted TPS and measuring false and true positive rates. In this experiment, we fixed the ground truth TPS cutoff to 1% since it is the most common and intuitive. **b** Comparing UIHC and single-cohort models across a range of 1% and the second cutoff value within the [2%, 75%] range, illustrating the 3-way classification accuracy. UIHC, universal immunohistochemistry model; H-Br, HER2 of breast; -Br, PD-L1 22C3 of breast; P-L, PD-L1 22C3 of lung; P-Bl, PD-L1 22C3 of bladder. AVG, average.

**Fig. 6. Histopathologic validation of the universal immunohistochemistry (UIHC) model. a** Lung cancer whole slide image (WSI) is stained with mesenchymal-epithelial transition factor (MET). The UIHC model predicts more accurate classes unlike the P-L model which confuses positively stained Tumor Cell (TC+) with negatively stained Tumor Cell (TC-). **b** Gastric cancer WSI is stained with fibroblast growth factor receptor 2 (FGFR2). P-L, PD-L1 22C3 of lung; TC, tumor cell.

**Fig. 7. Qualitative analysis of the artificial intelligence (AI)-learned representation**. **a** Two-dimensional (2D) projection of internal representation colored by tumor proportion score (TPS). Each patch is encoded to a 2D plot using three representations: raw pixels, self-supervised

learning model (SSL), and the universal immunohistochemistry (UIHC) model. Each dot represents one image patch from either an observed cohort available during training or from a novel cohort never seen by the UIHC model. The color represents the TPS within the patch. **b** A mosaic of image patches sorted by the internal representation. Using the same 2D representation as **a**, actual patches are displayed. **c** The assessment of cohort similarity through *p*-values. A higher *p*-value in UIHC signifies an inability to differentiate cohorts by UIHC, thus demonstrating the independence of UIHC from cohort effects.

**Fig. 8. Data pipeline for Universal Immunohistochemistry (UIHC) artificial intelligence (AI) model. a** Example of annotation process; patches extracted from whole slide images (WSIs), then cells are manually annotated by expert pathologists. WSIs are split into 0.04 mm$^2$ patches (resized to 1024×1024 pixels at 0.19 microns-per pixel). **b** Patch-level annotation count by its positivity (negatively stained Tumor Cell [TC-] or positively stained Tumor Cell [TC+]). **c** The number of WSI in the WSI-level dataset only for testing.

**Table 1. Validation of universal immunohistochemistry (UIHC) model on cases with next-generation sequencing results.**

| Case no. | Organ | Group | Mutation/Amplification detail | UIHC TPS | Average UIHC TPS according to the group |
|---|---|---|---|---|---|
| 1 | Lung | EGFR exon20ins | p.Ala763_Tyr764insPheGlnGluAla | 68.6 | 75.7±23.2 |
| 2 | Lung | EGFR exon20ins | p.Ala767_Val769dup | 85.7 | |
| 3 | Lung | EGFR exon20ins | p.Asp770_Asn771insGly | 88.4 | |
| 4 | Lung | EGFR exon20ins | p.Ser768_Asp770dup | 27.1 | |
| 5 | Lung | EGFR exon20ins | p.His773_Val774insThrHis | 80.0 | |
| 6 | Lung | EGFR exon20ins | p.Pro772_His773insProAsnPro | 98.0 | |
| 7 | Lung | EGFR exon20ins | p.P772_H773dup | 82.2 | |
| 8 | Lung | MET exon 14 skipping | c.3082+2T>G | 74.1 | 77.1±17.7 |
| 9 | Lung | MET exon 14 skipping | c.2942-28_2944del | 88.9 | |
| 10 | Lung | MET exon 14 skipping | c.3025C>T | 89.9 | |
| 11 | Lung | MET exon 14 skipping | c.3082+1G>C | 53.1 | |
| 12 | Lung | MET exon 14 skipping | c.3082+2T>C | 60.0 | |
| 13 | Lung | MET exon 14 skipping | c.3082G>T | 96.7 | |
| 14 | Lymph node | MET amplification | 8 copies | 94.6 | 94.5±2.0 |
| 15 | Lung | MET amplification | 4 copies | 92.5 | |

| 16 | Lung | MET amplification | 5 copies | 96.5 | |

EGFR, epidermal growth factor receptor; MET, mesenchymal-epithelial transition factor; TPS, tumor proportion score.

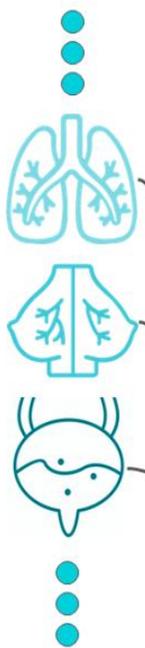
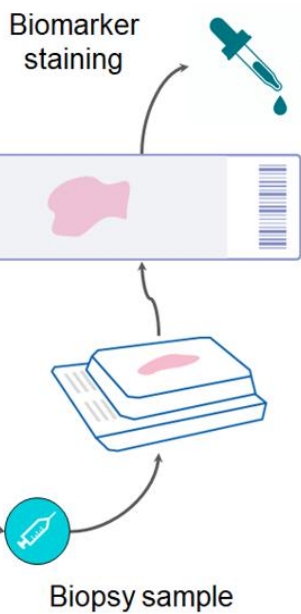
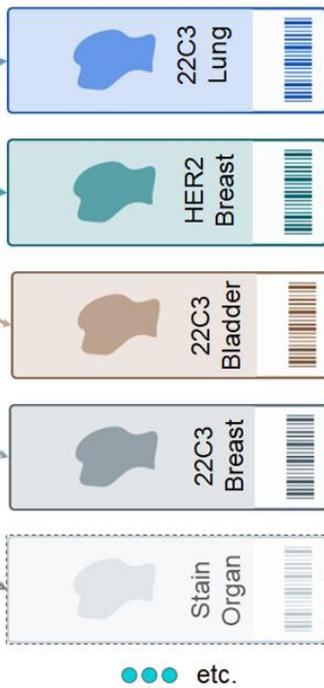
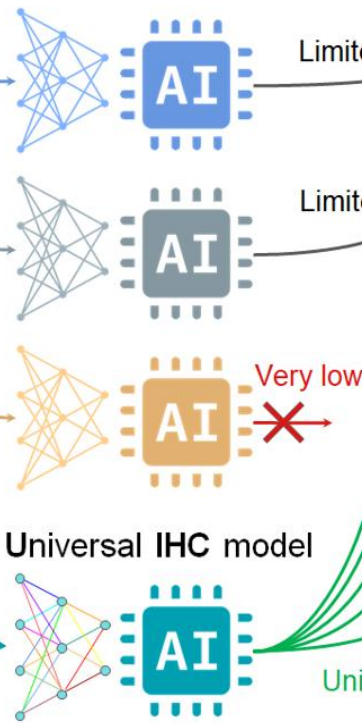
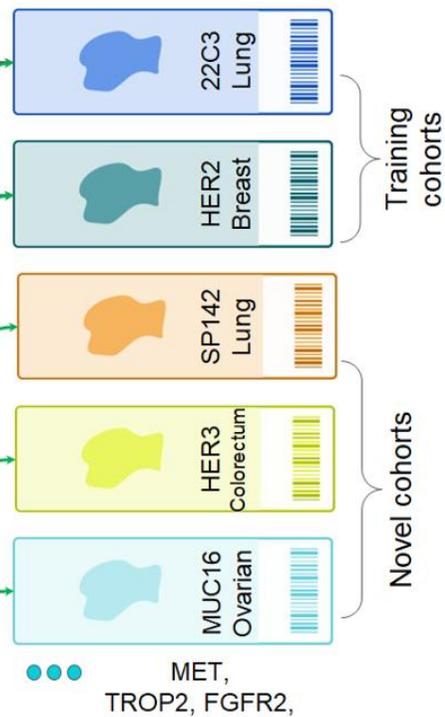

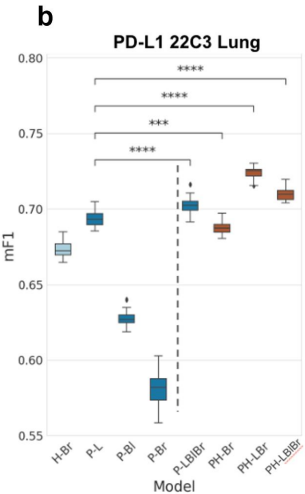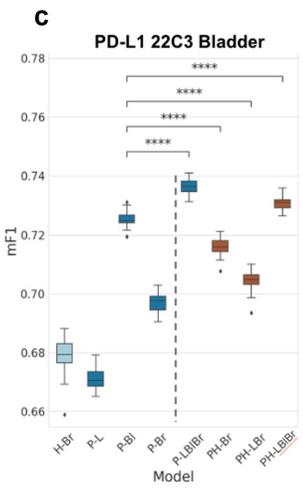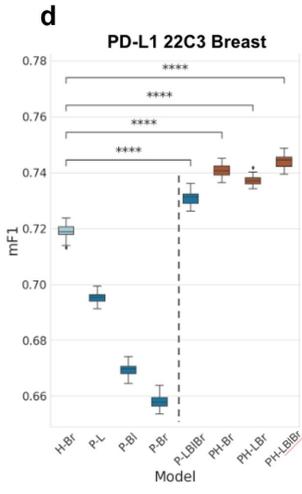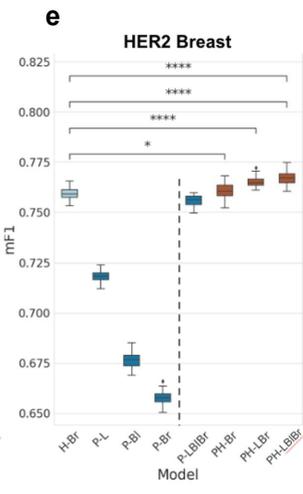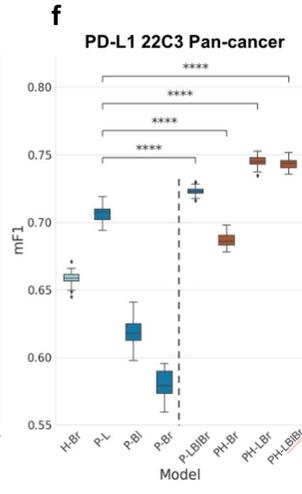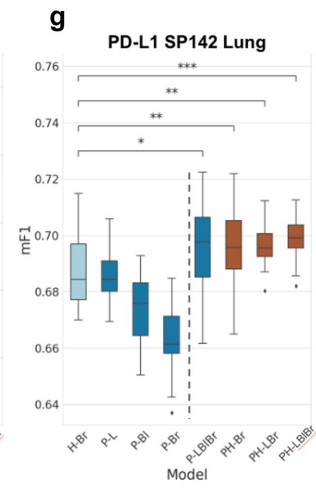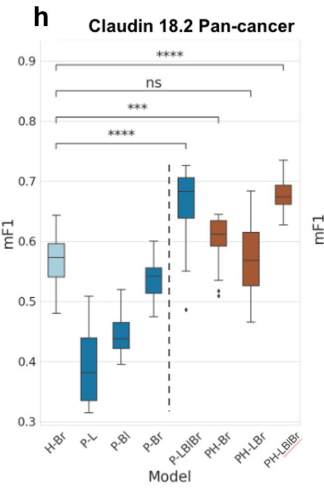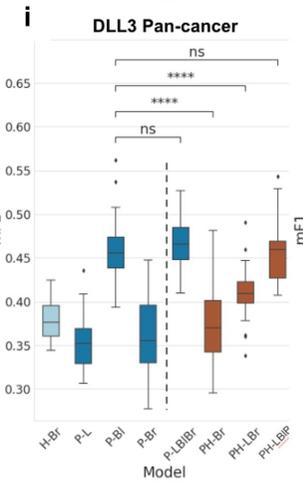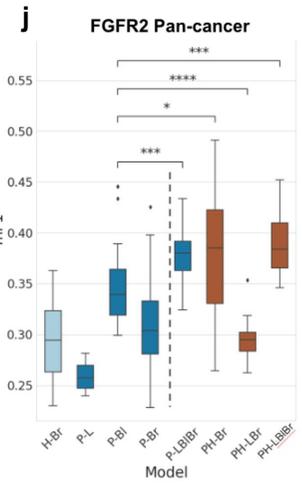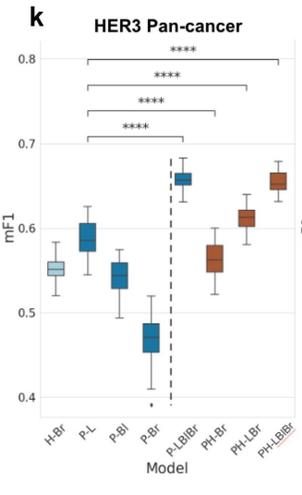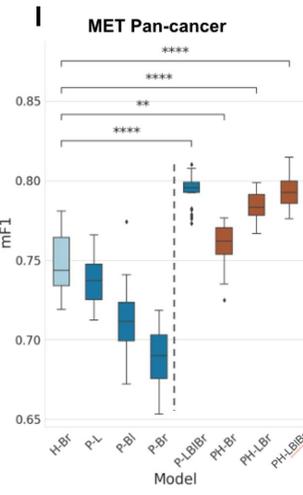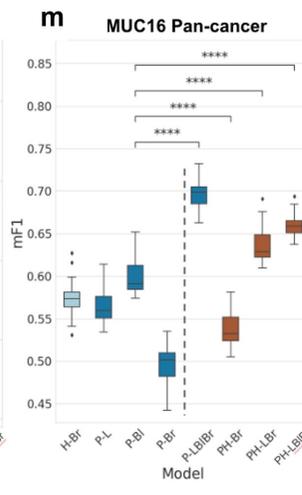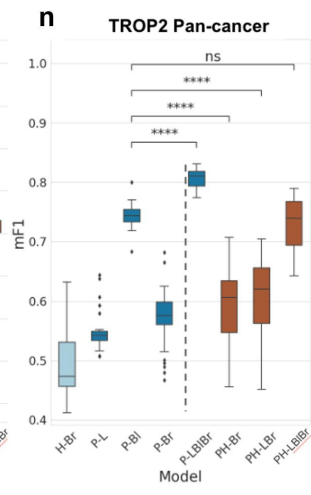

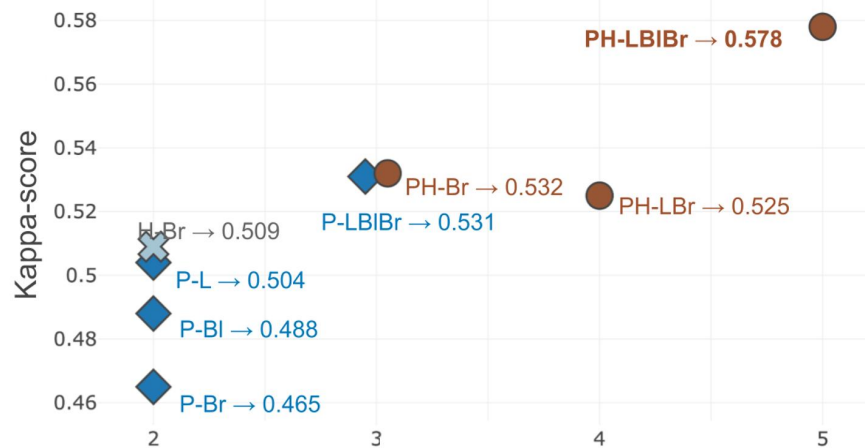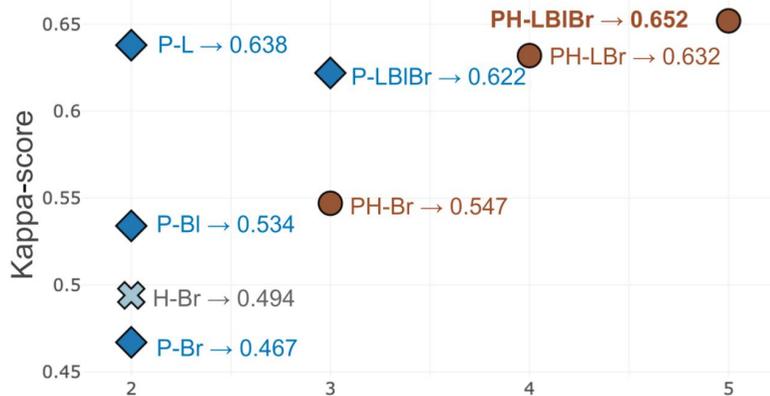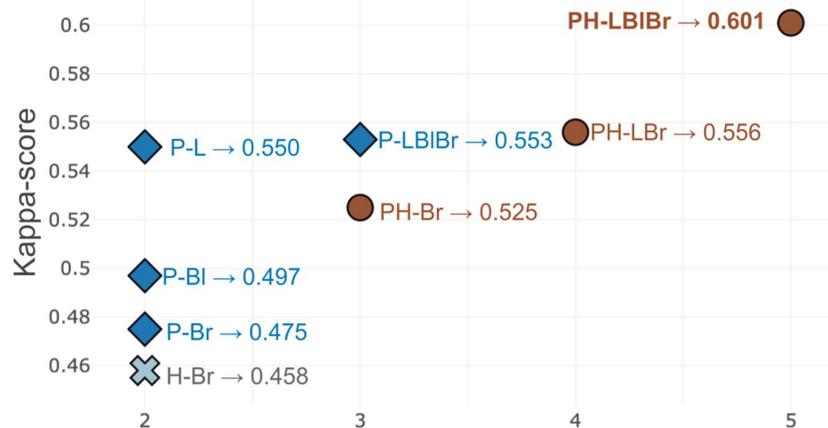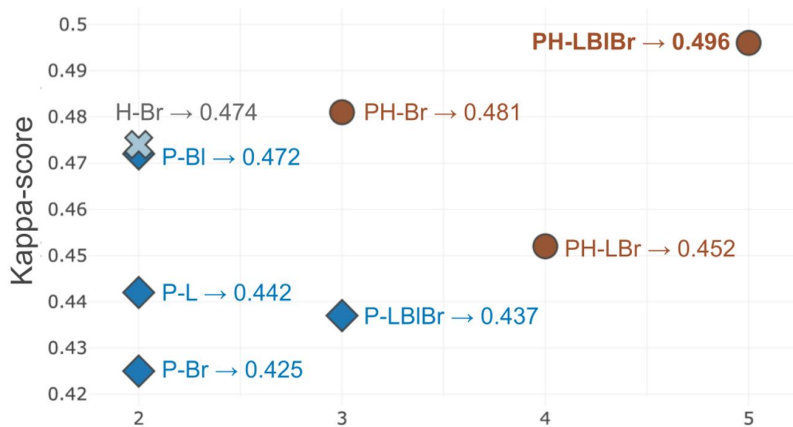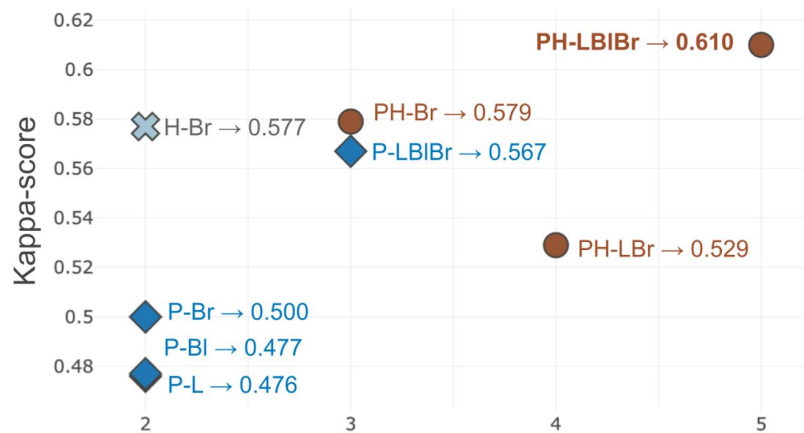

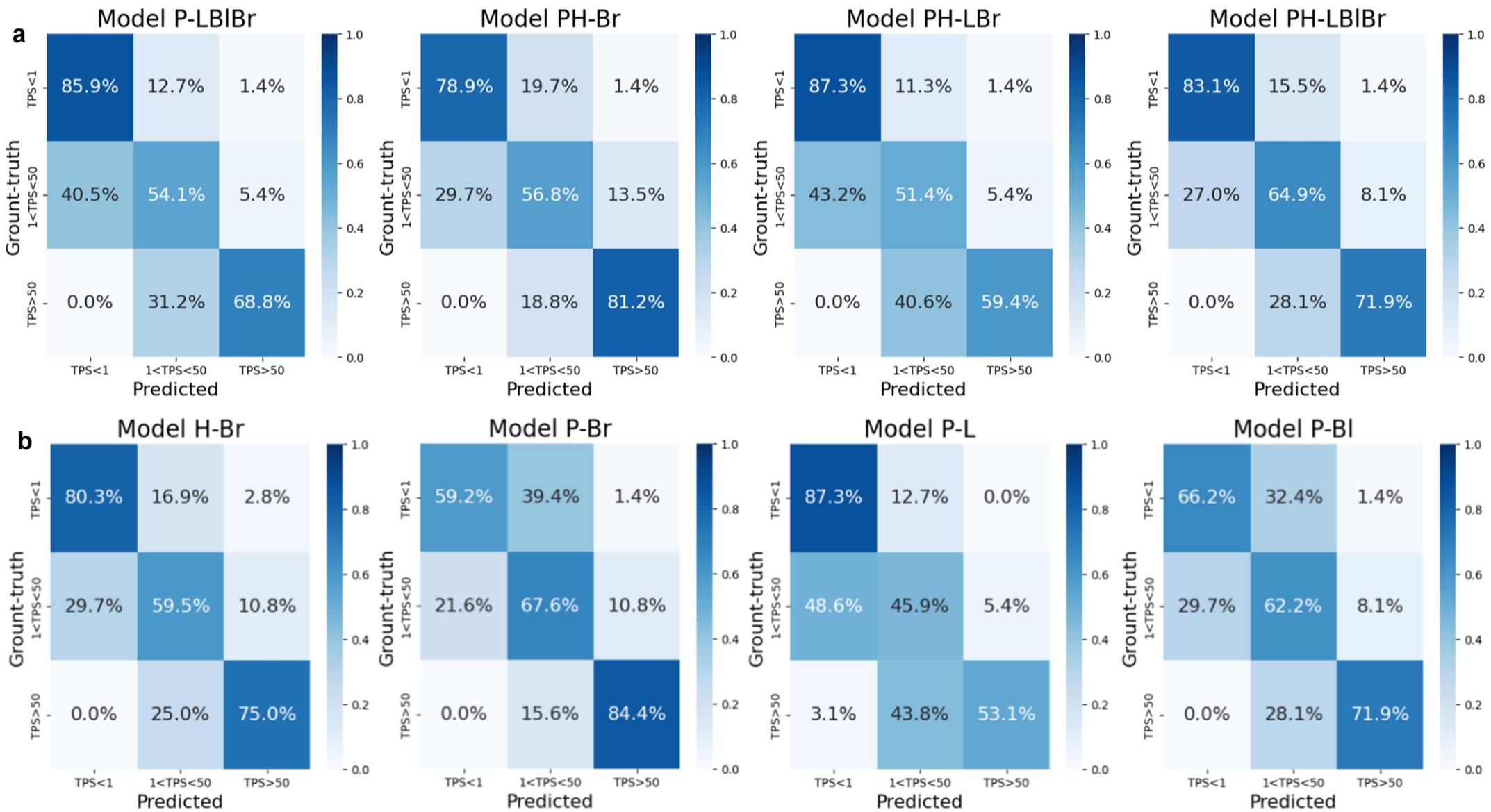

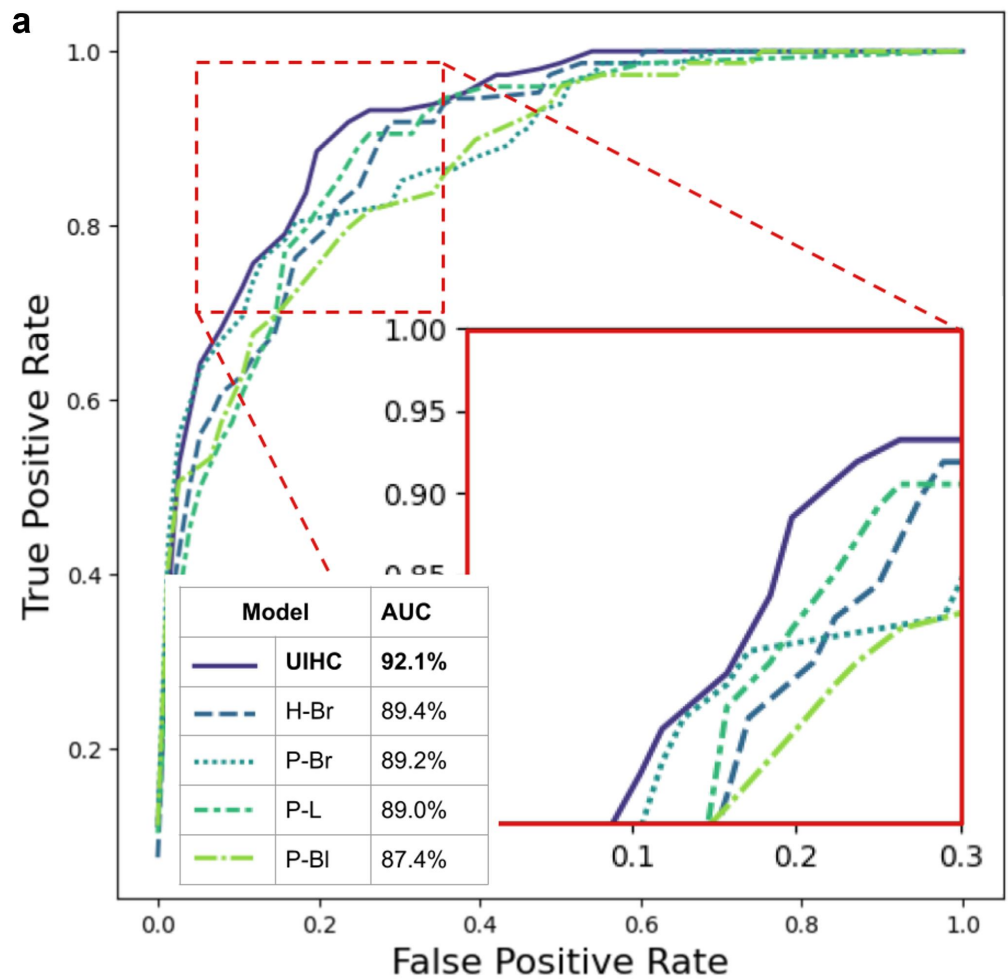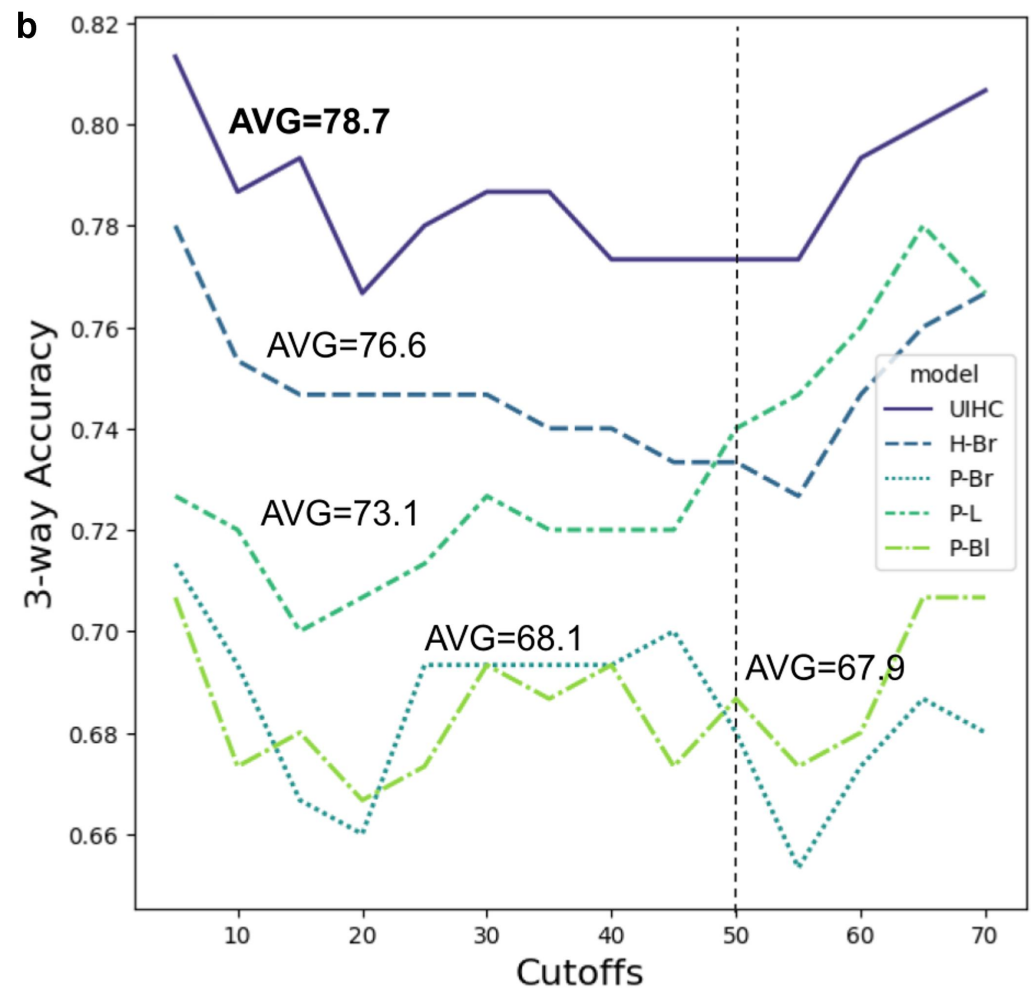

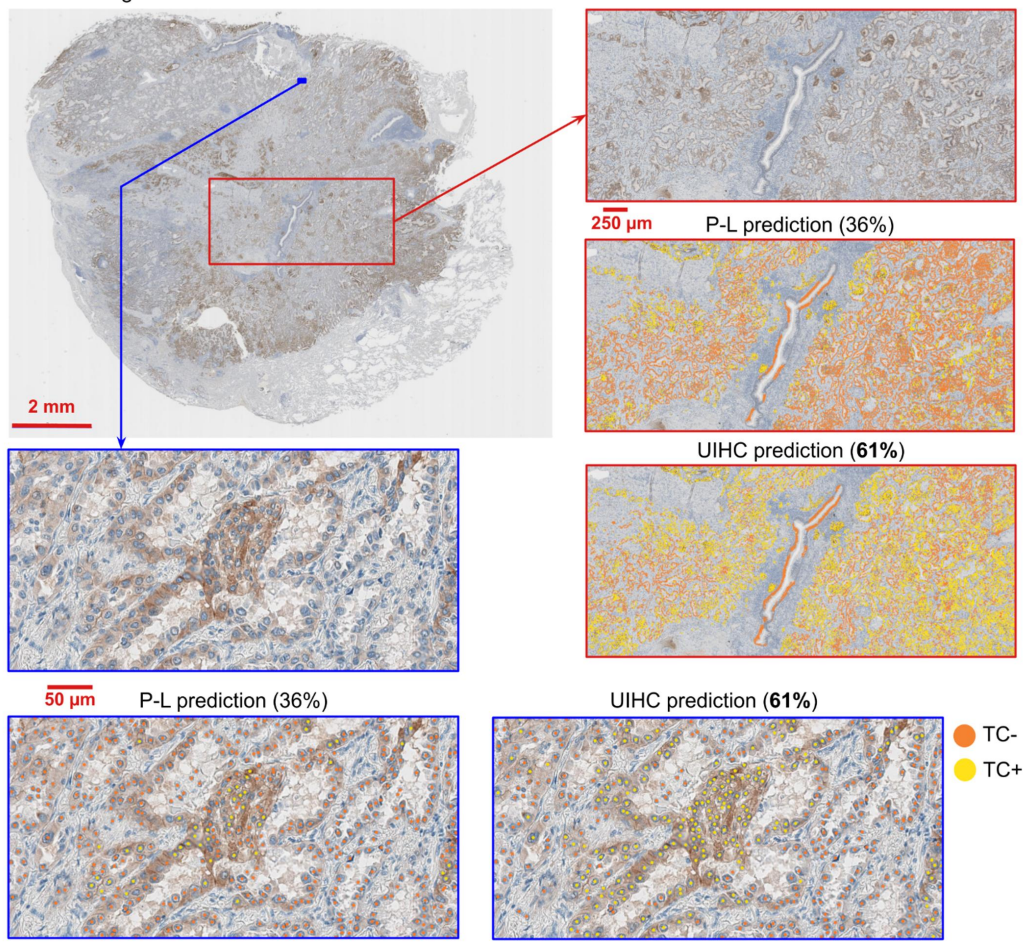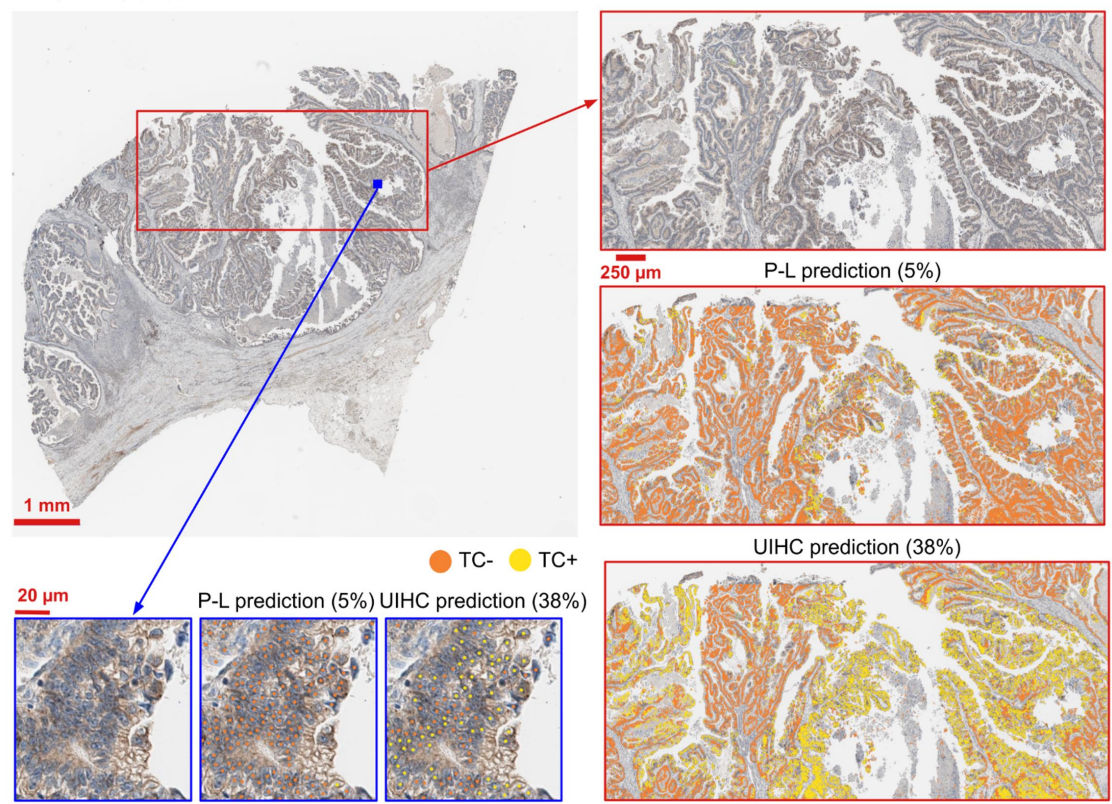

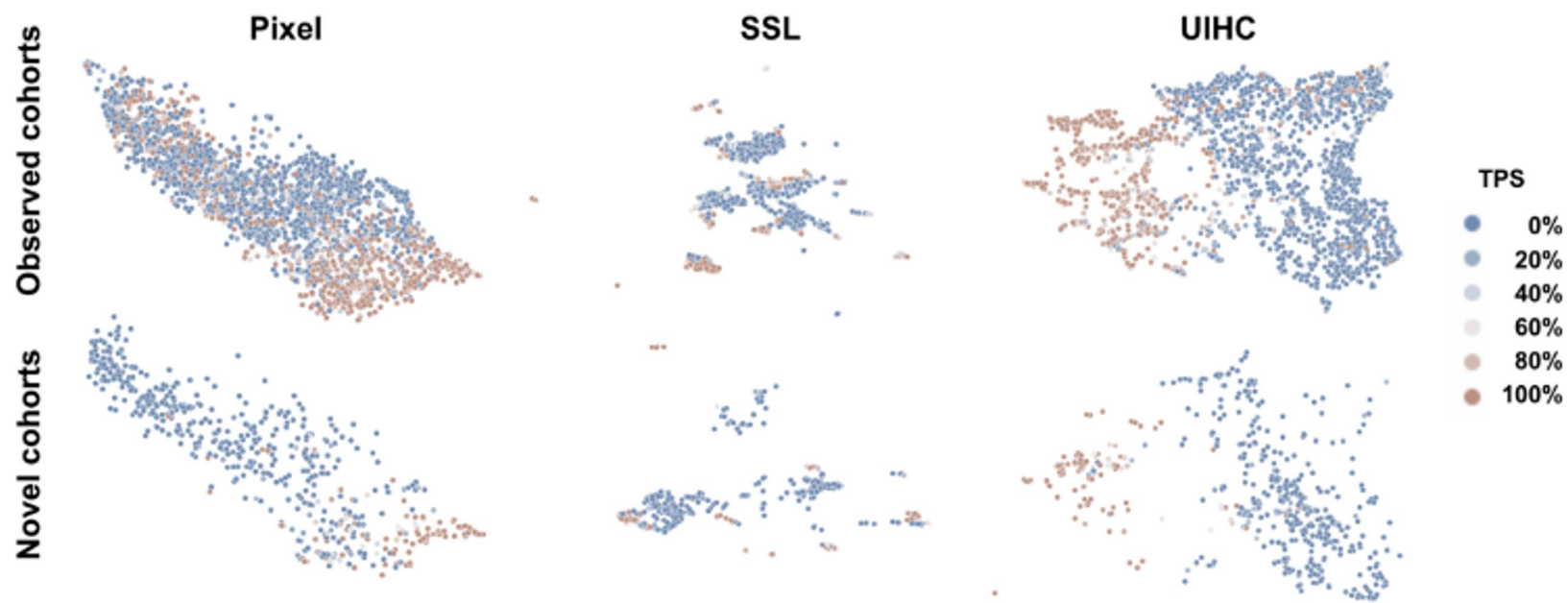
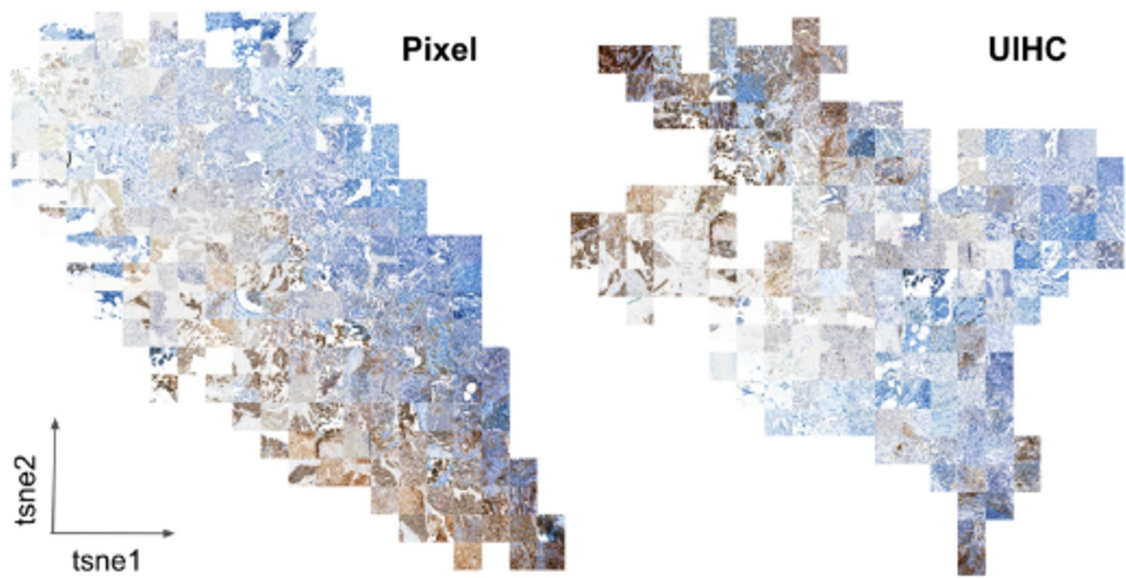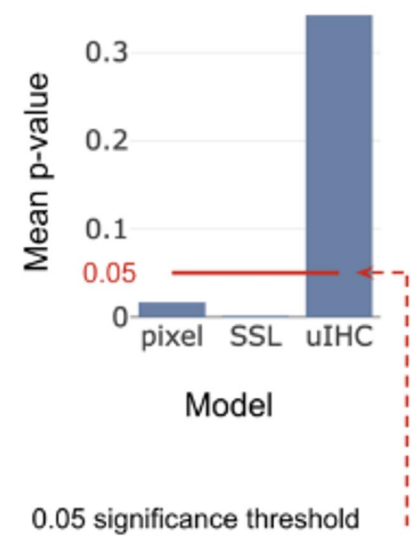

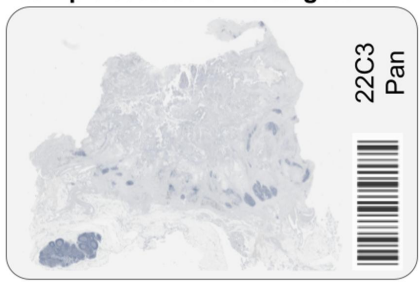
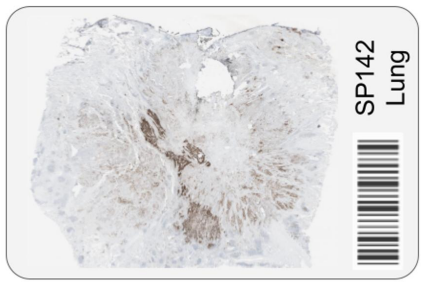
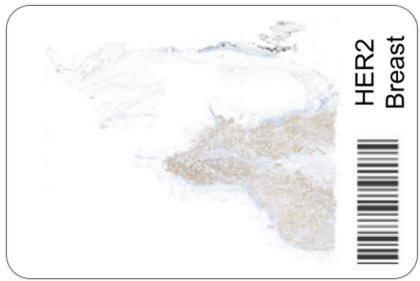
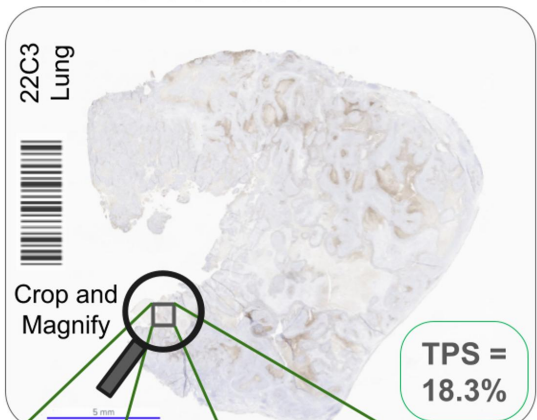
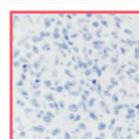
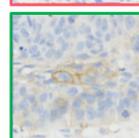
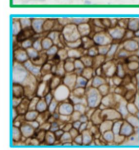
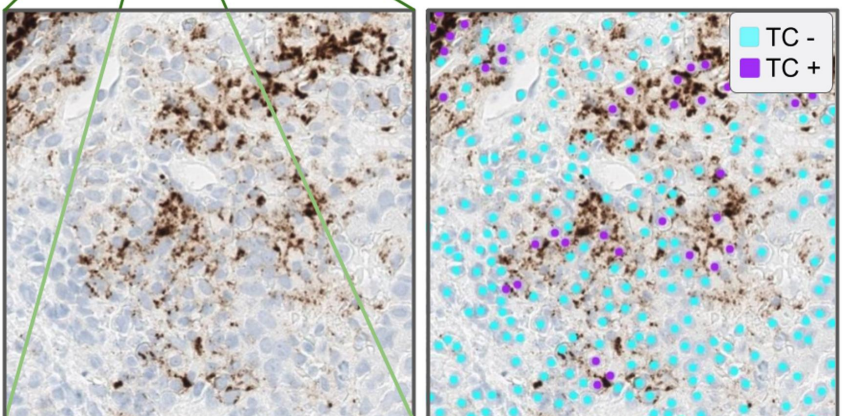
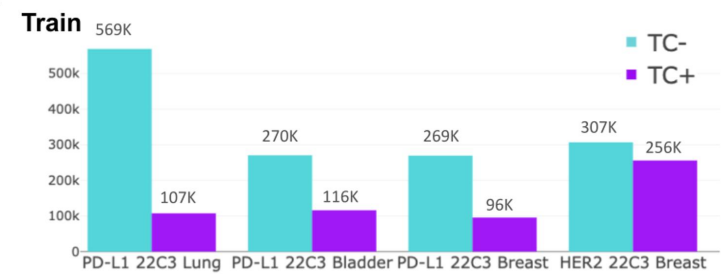
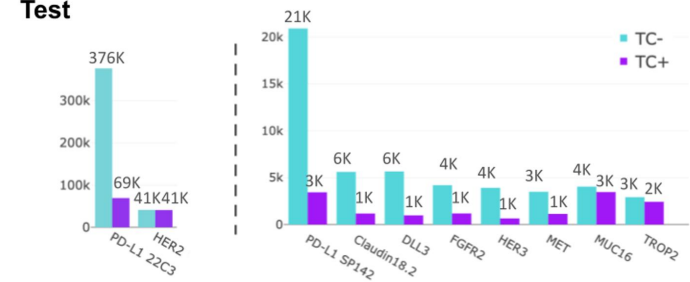
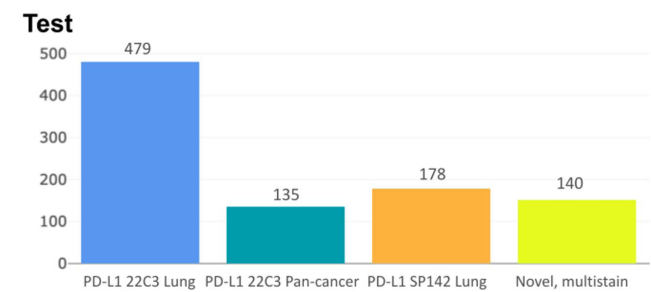

Supplementary Materials for

**Generalizing AI-driven Assessment of Immunohistochemistry across Immunostains and Cancer Types: A Universal Immunohistochemistry Analyzer**

Biagio Brattoli[1,+], Mohammad Mostafavi[1,+], Taebum Lee[1,+], Wonkyung Jung[1], Jeongun Ryu[1], Seonwook Park[1], Jongchan Park[1], Sergio Pereira[1], Seunghwan Shin[1], Sangjoon Choi[2], Hyojin Kim[3], Donggeun Yoo[1], Siraj M. Ali[1], Kyunghyun Paeng[1], Chan-Young Ock[1], Soo Ick Cho[1,\*], and Seokhwi Kim[4,5\*]

[+]These authors contributed equally to this work.

[*]Corresponding authors: Seokhwi Kim (seokhwikim@ajou.ac.kr), Soo Ick Cho (sooickcho@lunit.io)

**This file includes:**

Supplementary Methods

Supplementary Figs. 1 to 4

Supplementary Tables 1 to 5

References (1 to 4)

**Supplementary Methods**

Quantitative evaluation of F1 score at patch-level

At the patch level, the F1 score is used as an evaluation metric for cell detection. The F1 score is a popular metric for object detection in computer vision since it considers precision and sensitivity simultaneously. Its definition requires the number of true positives (TP), false positives (FP), and false negatives (FN).

$$F1\ score = 2 \times \frac{precision \times recall}{(precision + recall)} = \frac{\#TP}{\#TP + 0.5 \times (\#FP + \#FN)}$$

Since these metrics have been developed for the classification task, it is not obvious how to measure them in the context of object detection. Therefore, a hit criterion is defined as follows. For each cell class, we determine the TP, FP, and FN with the following process,

1. Sort cell predictions by their confidence score.

2. Starting from a cell prediction with the highest confidence score, check whether any ground-truth cell is within a valid distance (25 pixels in 0.19 microns per pixel (MPP)) from the cell prediction.

   2-1. If there is no ground-truth cell within a valid distance, the cell prediction is counted as an FP.

   2-2. If there are one or more ground-truth cells within a valid distance, the cell prediction is counted as TP. The nearest ground-truth cell is matched with the cell prediction and ignored from the further process.

3. Go back to 2. until the cell prediction with the lowest confidence score is reached.

4. The remaining ground-truth cells that are not matched with any cell prediction are counted

as FN.

After the above process, we aggregate the total number of TP, FP, and FN per cell class over all samples; then, we compute the per-class F1 score. The mean F1 (mF1) score across cell classes (TC- and TC+) is used as the final score. Each result has been reproduced ×30 times using Monte-Carlo dropout at test time, an algorithm developed to study the robustness of a deep learning model over small variations[1]. In Fig. 1, for each test set we compare the statistical significance (p-value < 0.05) between the best single-cohort model (left of the dotted line) with all other multiple-cohort models. *p*-value has been calculated using the Wilcoxon signed-rank test implemented by *scipy.stats*[2,3].

Quantitative evaluation of tumor proportion score at slide level

Tumor proportion score (TPS) is calculated by the following equation:

$$TPS = 100 \times \frac{\#TC+}{\#TC- + \#TC+}$$

Programmed Death-Ligand 1 (PD-L1) expression is subgrouped according to the TPS cutoff 1% and 50%, i.e., classified into TPS < 1%, 1% ≤ TPS < 50%, and TPS ≥ 50%. For simplicity, TPS has been used as a general whole slide image (WSI)-level metric to compare all IHC quantification across AI models. For each WSI, three board-certified pathologists assign a TPS score following the official protocol[4]. A category (TPS < 1%, 1% ≤ TPS < 50%, or TPS ≥ 50%) is then assigned to the slide by applying the cutoff, for example for PD-L1 Lung. The WSI-level ground truth was based on the consensus of three board-certified pathologists. For example, if two pathologists determined a WSI as TPS<1% while one pathologist determined it as TPS 1-

49%, the WSI was assigned to TPS<1%.

As explained above, we compute TPS using our model for each WSI, and compare it with the manually assigned one. Fig. 2 shows the standard accuracy computed as follows, given *N* number of WSIs in a test set (e.g., HER2 Breast), *y* and *ŷ* are respectively ground truth (GT) and predicted category.

$$Accuracy = \frac{1}{N}\sum_{i=1}^{N} [y_i = \hat{y}_i]$$

**Supplementary Fig. 1. Patch-level quantitative analysis of the artificial intelligence (AI) models sorted by cancer types.** H-Br, HER2 of breast; P-L, PD-L1 22C3 of lung; P-Br, PD-L1 22C3 of breast; P-LBlBr, PD-L1 22C3 of lung, bladder, and breast; PH-B, PD-L1 22C3 and HER2 of breast; PH-LBr, PD-L1 22C3 and HER2 of lung and breast; PH-LBlBr, PD-L1 22C3 and HER2 of lung, bladder, and breast.

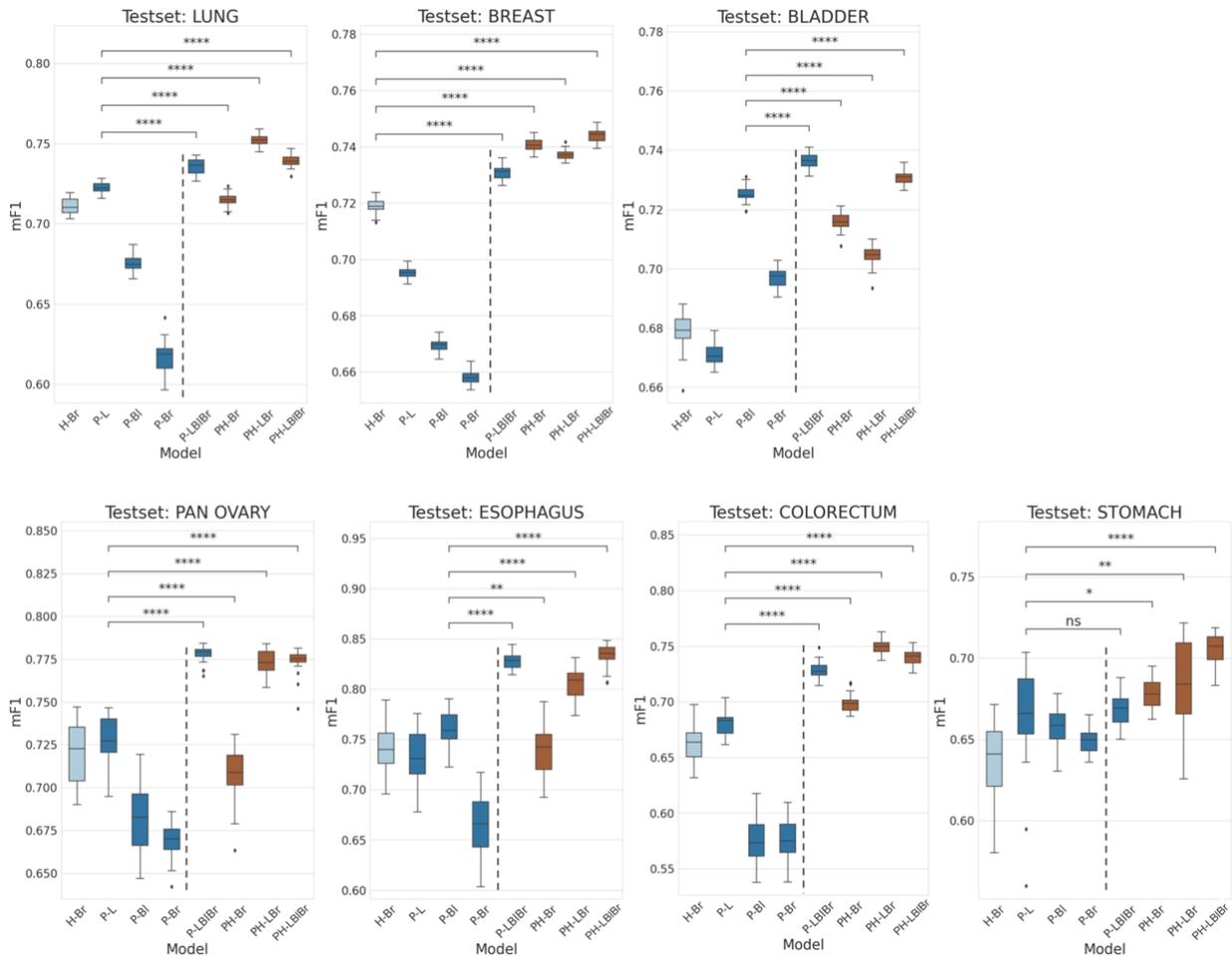

7   **Supplementary Fig. 2. Whole slide image (WSI)-level accuracy analysis of the artificial**
8   **intelligence (AI) models. a** Macro-averaged accuracy of the eight AI models over all the stains.
9   **b** Accuracy of the AI models in PD-L1 22C3 Lung dataset. **c** Accuracy of the AI models in PD-
10  L1 22C3 Pan-cancer dataset. **d** Accuracy of the AI models in PD-L1 SP142 Lung dataset. **e**
11  Accuracy of the AI models in multi-stain Pan-cancer dataset. The X-axis presents the summation
12  of utilized stain types and the organ types of each cohort when training. Accuracy metrics are
13  presented for 3-classes whole slide image (WSI) evaluation based on tumor proportion score
14  (TPS) cutoffs. H-Br, HER2 of breast; P-L, PD-L1 22C3 of lung; P-Br, PD-L1 22C3 of breast; P-
15  LBlBr, PD-L1 22C3 of lung, bladder, and breast; PH-B, PD-L1 22C3 and HER2 of breast; PH-
16  LBr, PD-L1 22C3 and HER2 of lung and breast; PH-LBlBr, PD-L1 22C3 and HER2 of lung,
17  bladder, and breast.

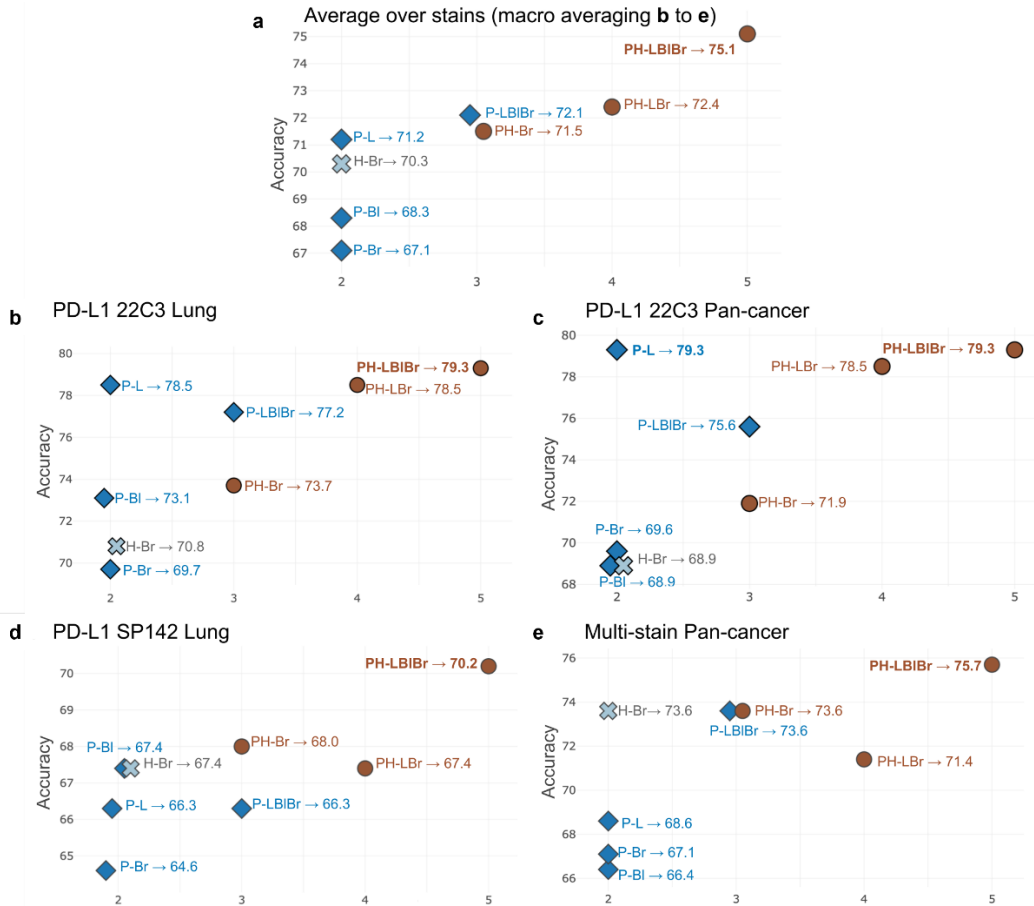

**Supplementary Fig. 3. Examples of whole slide images (WSIs) and their tumor proportion score (TPS) from training cohorts. a** Samples from training cohorts. **b** Samples from novel cohorts.

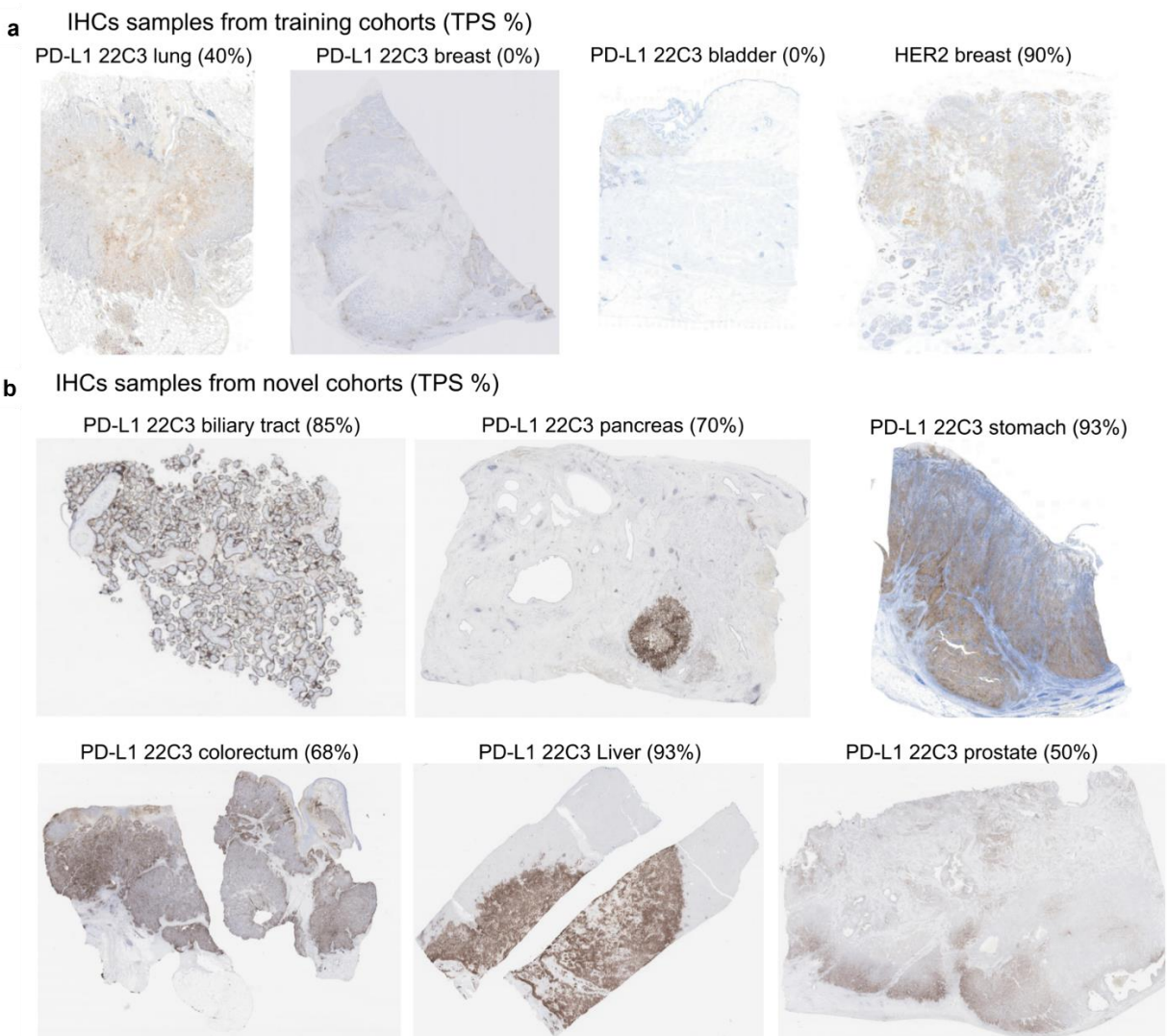

25   **Supplementary Fig. 4. Examples of whole slide images (WSIs) and their tumor proportion**
26   **score (TPS) from novel cohorts.**

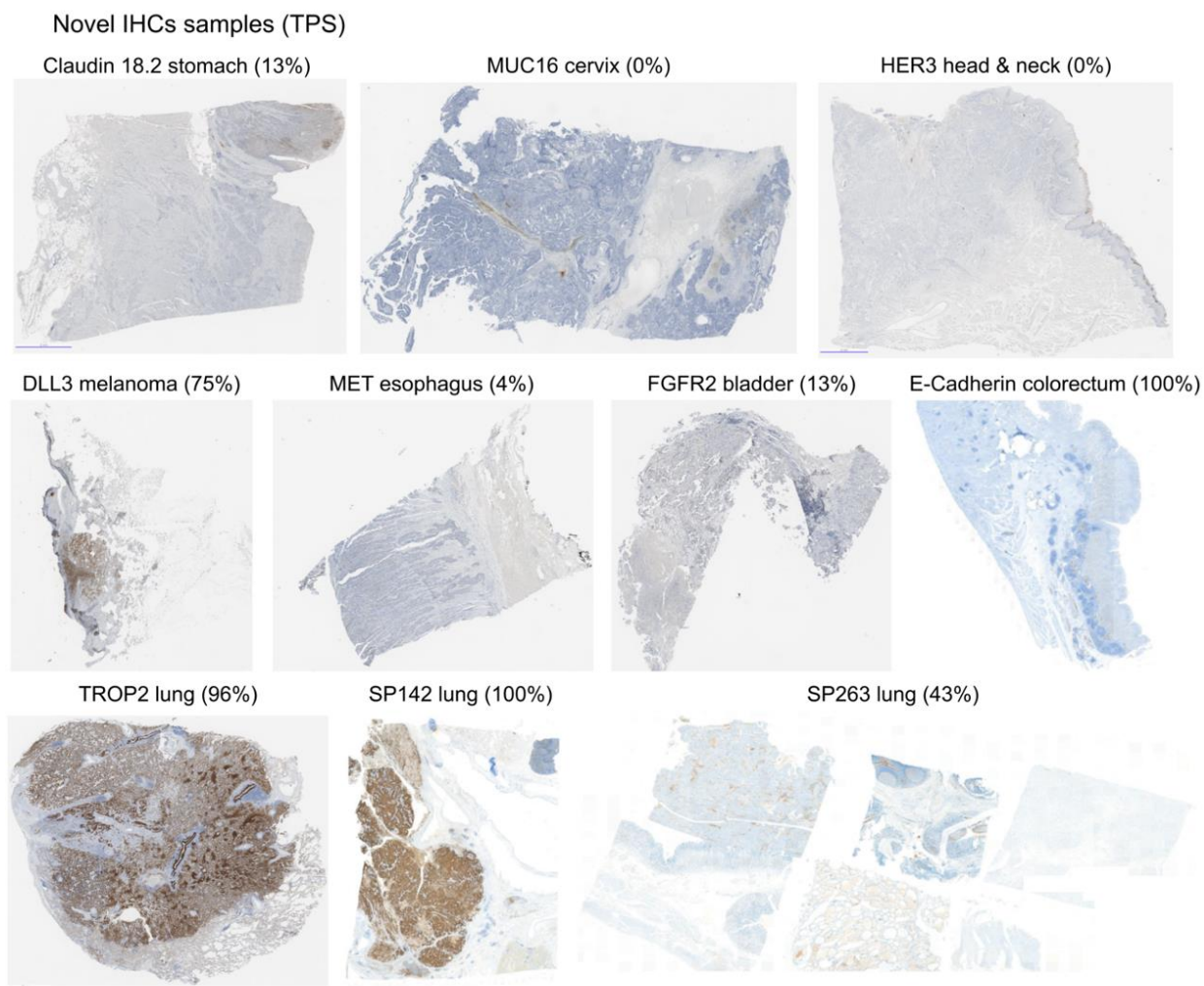

27
28

29  **Supplementary Table 1. Number of slides for each whole slide image (WSI) test sets,**

30  **divided by 1%/50% tumor proportion score (TPS) cutoffs into three classes.**

| Dataset | TPS < 1% | TPS 1 − 49% | TPS ≥ 50% |
|---|---|---|---|
| PD-L1 22C3 Lung (n = 479) | 81 (16.9%) | 162 (33.8%) | 236 (49.3%) |
| PD-L1 22C3 Pan-cancer (n = 135) | 95 (70.4%) | 25 (18.5%) | 15 (11.1%) |
| PD-L1 SP142 Lung (n = 178) | 81 (45.5%) | 71 (39.9%) | 26 (14.6%) |
| Novel, multi-stain (n = 140) | 71 (50.7%) | 37 (26.4%) | 32 (22.9%) |
| Total (n = 932) | 328 (35.2%) | 295 (31.7%) | 309 (33.2%) |

32  **Supplementary Table 2. Dataset configuration of the whole slide images (WSIs) used to**

33  **develop the universal immunohistochemistry (UIHC) model.**

| Dataset | Source | Training | Tuning | Internal test | Total |
|---|---|---|---|---|---|
| PD-L1 22C3 lung | Cureline | 329 | 41 | 0 | 370 |
| | Aurora Dx | 381 | 32 | 50 | 463 |
| | SNUBH | 0 | 28 | 29 | 57 |
| | SMC | 0 | 62 | 49 | 111 |
| PD-L1 22C3 urothelial carcinoma | Cureline | 281 | 79 | 40 | 400 |
| PD-L1 22C3 Breast cancer | Cureline | 281 | 83 | 40 | 404 |
| HER2 4B5 Breast cancer | Cureline | 392 | 119 | 58 | 569 |
| | AuroraDx | 415 | 118 | 60 | 593 |
| | Superbiochips | 60 | 13 | 6 | 79 |
| Total | | 2,139 | 575 | 332 | 3,046 |

34  PD-L1, Programmed Death-Ligand 1; HER2, Human Epidermal growth factor Receptor 2; SMC,
35  Samsung Medical Center; SNUBH, Seoul National University Bundang Hospital

36  **Supplementary Table 3. Antibody information for immunostains.**

| Target | Antibody | Manufacturer | Dilution factor | Stain localization |
|--------|----------|--------------|-----------------|--------------------|
| Claudin 18.2 | Anti-Claudin18.2 antibody [EPR19202] ab222512 | abcam | 1:50 | Membrane |
| DeLta-Like 3 (DLL3) | DLL3 (SP347), 08416931001 | Ventana | RTU | Membrane, Cytoplasm |
| E-cadherin | E-CAD-L-CE / Leica | Leica | 1:100 | Membrane |
| Fibroblast Growth Factor Receptor 2 (FGFR2) | Anti-FGFR2 antibody [SP273] – N-terminal ab227683 | abcam | 1:100 | Membrane, Cytoplasm |
| Human Epidermal growth factor Receptor 3 (HER3) | Anti-ErbB3/HER3 antibody [SP71] ab93739 | abcam | 1:100 | Membrane, Cytoplasm |
| Mesenchymal-Epithelial Transition factor (MET) | anti-Total c-MET (SP44) | Ventana | RTU | Membrane, Cytoplasm |
| MUCin-16 (MUC16) | CA-125 (OC125), 05267269001 | Cell Marque Corporation | RTU | Membrane |
| TROPhoblast cell-surface antigen 2 (TROP2) | Anti-TROP2 antibody [EPR20043] ab214488 | abcam | 1:1000 | Membrane |

38  **Supplementary Table 4. Dataset configuration at the independent tumor cell level used to**
39  **develop the universal immunohistochemistry (UIHC) model.**

| Dataset | Class | Train | Tune | Internal test | Total |
|---|---|---|---|---|---|
| PD-L1 22C3 lung | TC+ | 107,438 | 28,763 | 13,792 | 149,993 |
| | TC- | 568,814 | 59,271 | 37,716 | 665,801 |
| PD-L1 22C3 urothelial carcinoma | TC+ | 115,965 | 19,453 | 17,654 | 153,072 |
| | TC- | 270,251 | 87,915 | 44,106 | 402,272 |
| PD-L1 22C3 breast cancer | TC+ | 95,556 | 19,331 | 9,789 | 124,676 |
| | TC- | 269,289 | 71,733 | 38,382 | 379,404 |
| HER2 4B5 Breast cancer | TC+ | 255,661 | 70,882 | 40,805 | 367,348 |
| | TC- | 306,679 | 97,889 | 41,216 | 445,784 |

40  PD-L1, Programmed Death-Ligand 1; HER2, Human Epidermal growth factor Receptor 2; TC+,
41  positively stained tumor cell; TC-, negatively stained tumor cell

42  **Supplementary Table 5. Dataset configuration at the independent tumor cell level used to**

43  **test the universal immunohistochemistry (UIHC) model.**

|  | Positively stained tumor cell (TC+) | Negatively stained tumor cell (TC−) |
|---|---|---|
| PD-L1 22C3 lung | 13,792 | 37,716 |
| PD-L1 22C3 urothelial carcinoma | 17,654 | 44,106 |
| PD-L1 22C3 breast cancer | 9,789 | 38,382 |
| HER2 4B5 Breast cancer | 40,805 | 41,216 |
| PD-L1 22C3 pan-cancer (biliary tract, colorectum, liver, stomach, prostate, and pancreas) | 7,259 | 34,491 |
| PD-L1 SP142 lung | 3,437 | 20,916 |
| Claudin 18.2 | 1,173 | 5,614 |
| MET | 1,110 | 5,614 |
| TROP2 | 2,423 | 2,917 |
| MUC16 | 3,464 | 4,043 |
| DLL3 | 974 | 5,664 |
| FGFR2 | 1,181 | 4,193 |
| HER3 | 650 | 3,916 |

44  PD-L1, Programmed Death-Ligand 1; HER2, Human Epidermal growth factor Receptor 2; MET,
45  Mesenchymal-epithelial transition factor; TROP2, Trophoblast cell-surface antigen 2; DLL3,
46  Delta-like 3; FGFR2, Fibroblast growth factor receptor 2; HER3, Human epidermal growth
47  factor receptor 3.

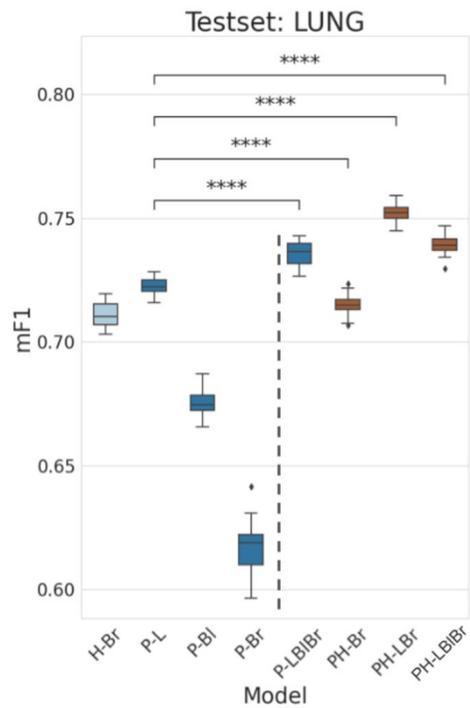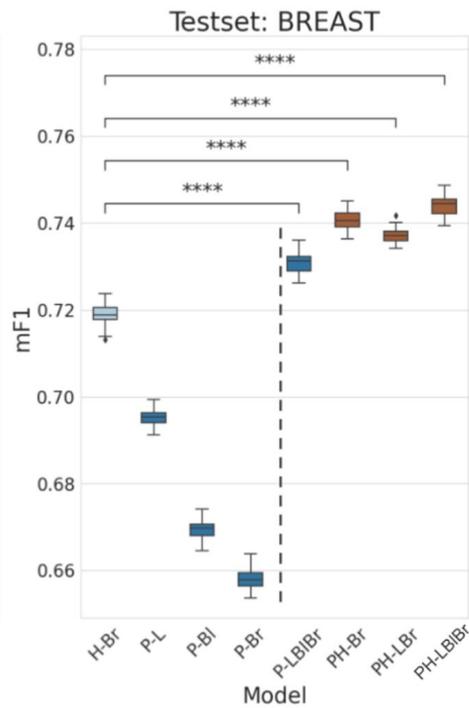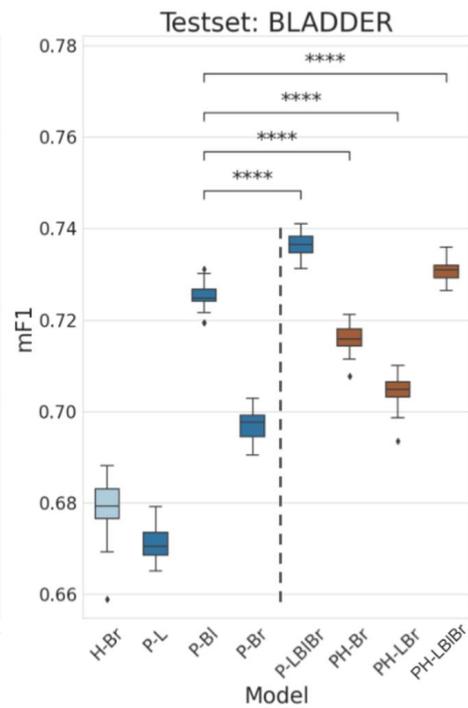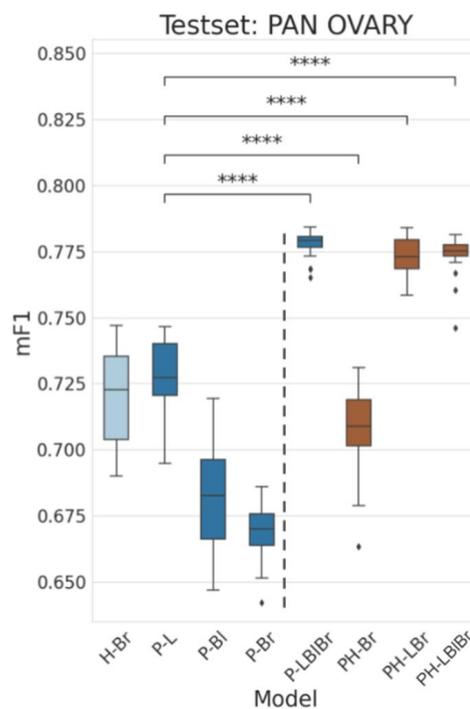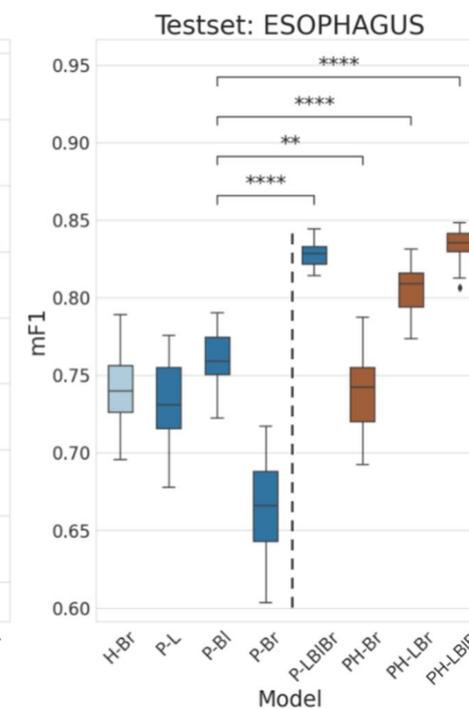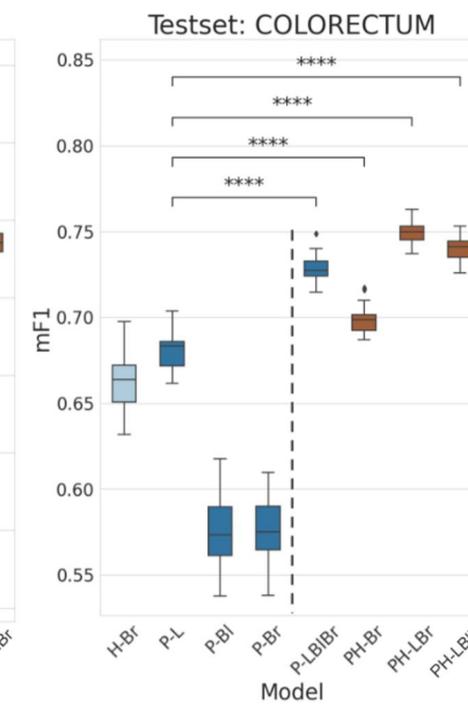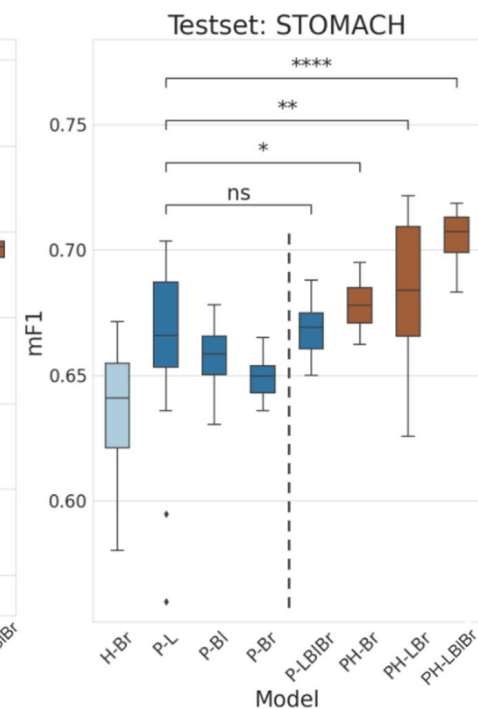

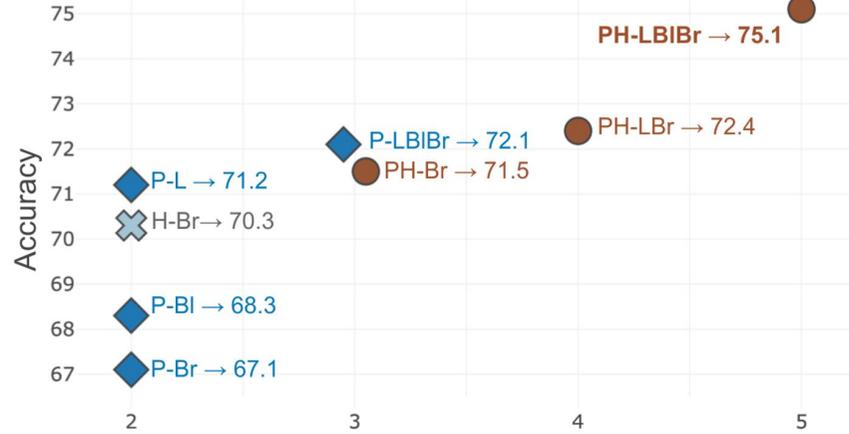
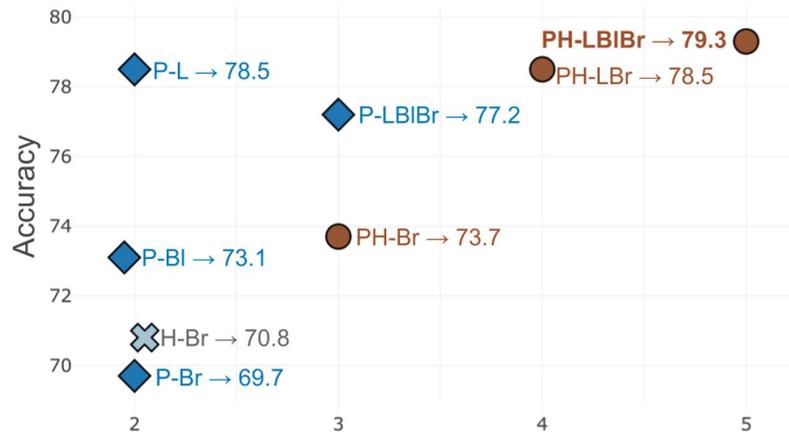
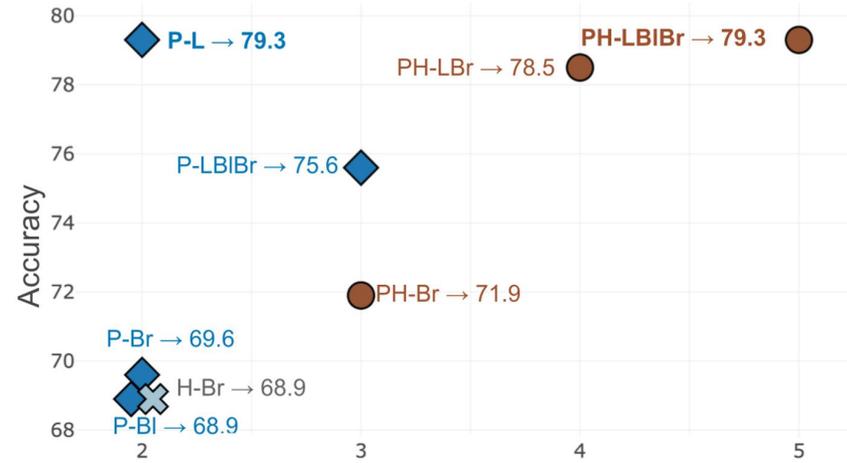
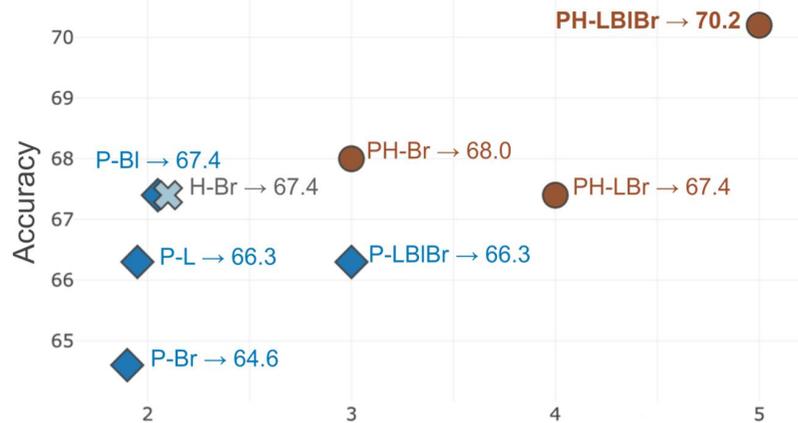
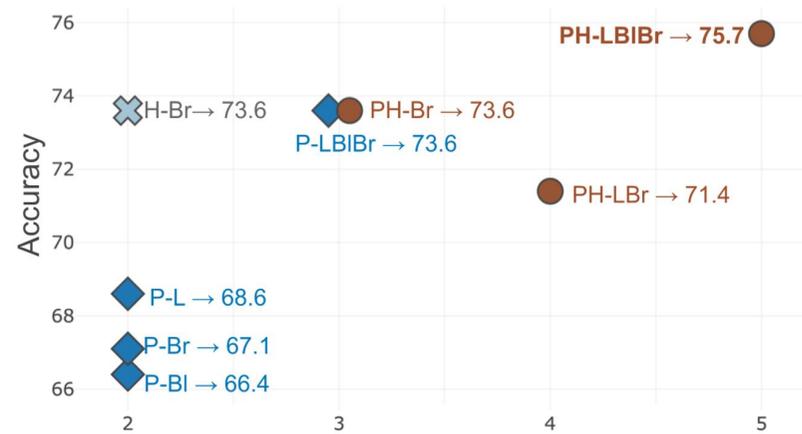

**a** IHCs samples from training cohorts (TPS %)

PD-L1 22C3 lung (40%)    PD-L1 22C3 breast (0%)    PD-L1 22C3 bladder (0%)    HER2 breast (90%)

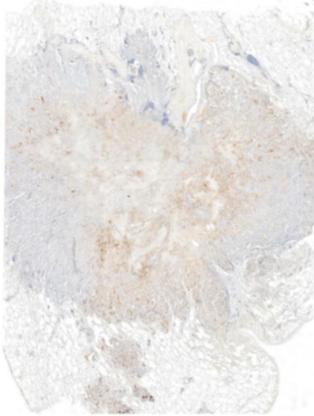 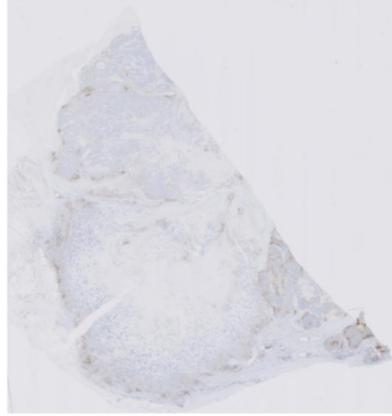 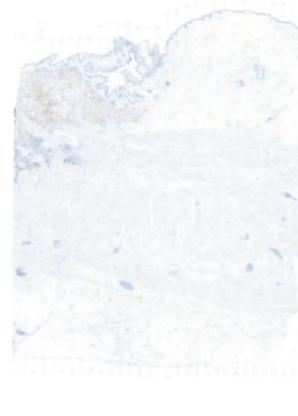 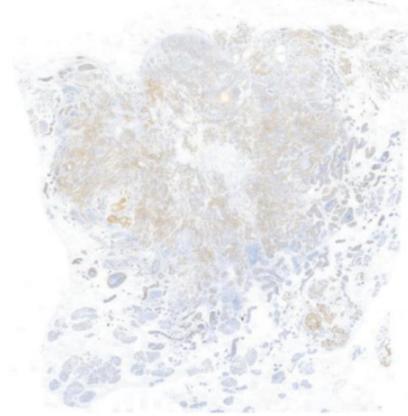

**b** IHCs samples from novel cohorts (TPS %)

PD-L1 22C3 biliary tract (85%)    PD-L1 22C3 pancreas (70%)    PD-L1 22C3 stomach (93%)

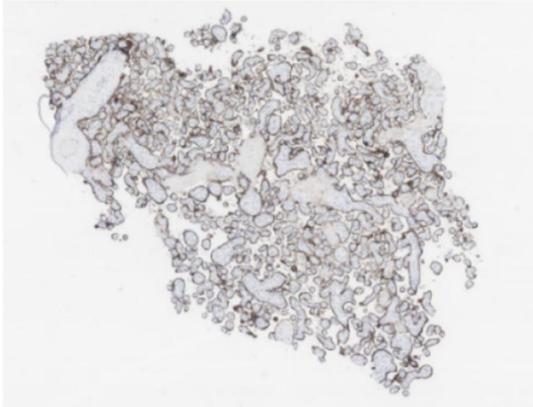 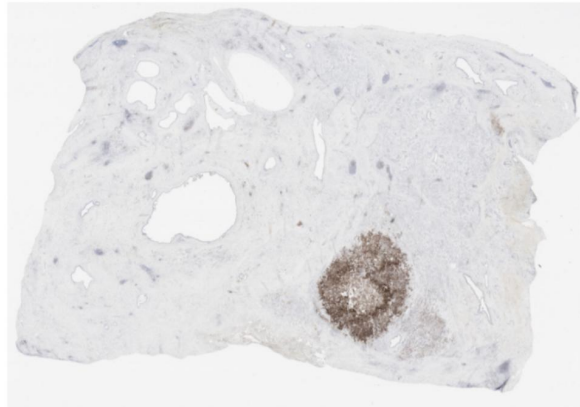 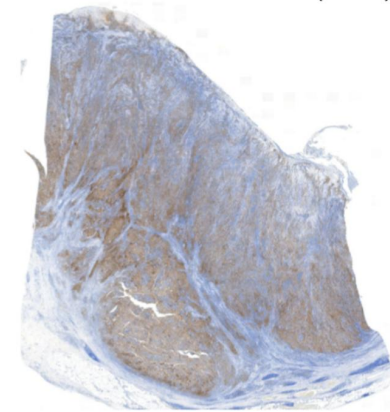

PD-L1 22C3 colorectum (68%)    PD-L1 22C3 Liver (93%)    PD-L1 22C3 prostate (50%)

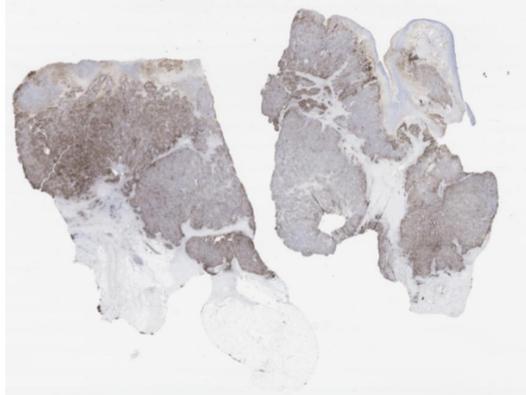 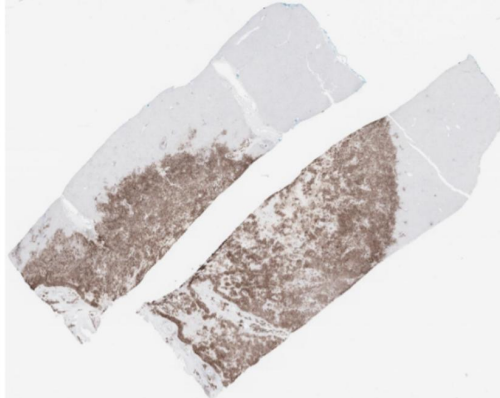 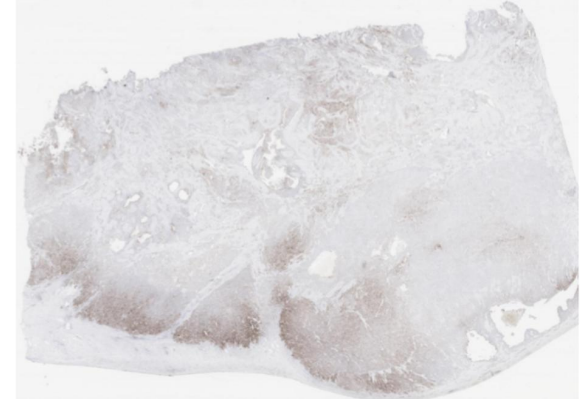

## Novel IHCs samples (TPS)

Claudin 18.2 stomach (13%)
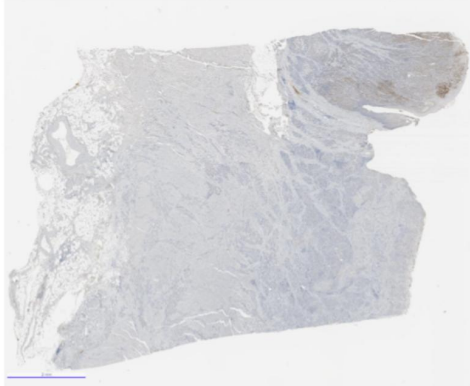

MUC16 cervix (0%)
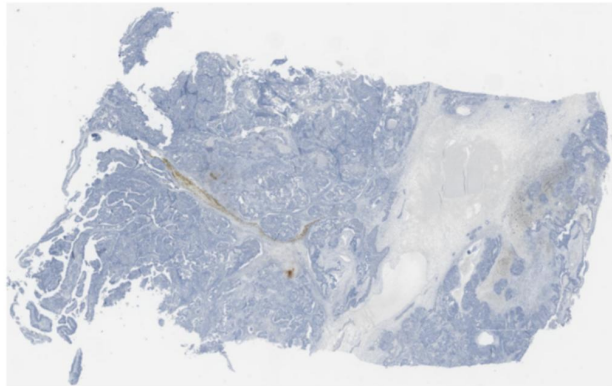

HER3 head & neck (0%)
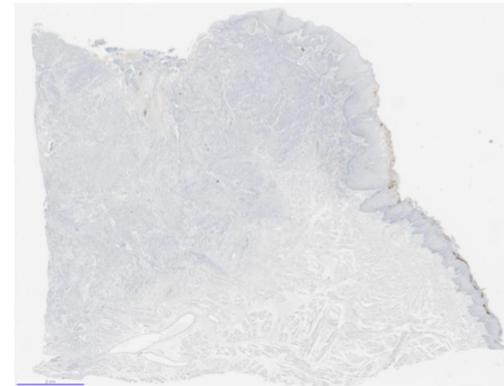

DLL3 melanoma (75%)
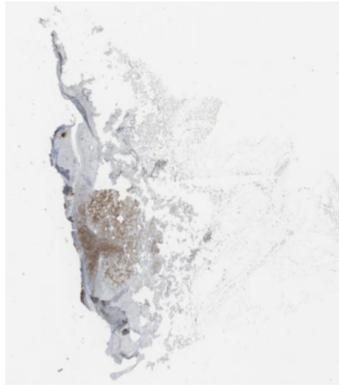

MET esophagus (4%)
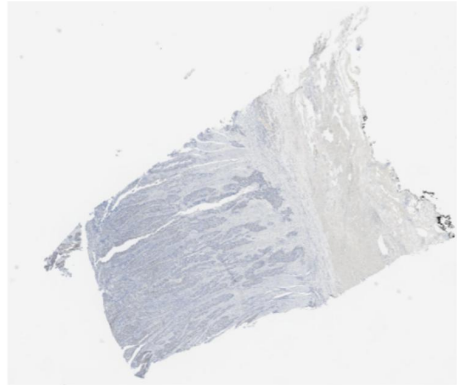

FGFR2 bladder (13%)
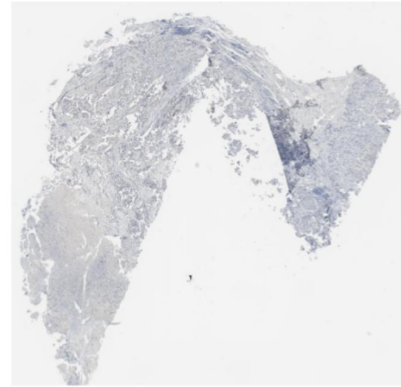

E-Cadherin colorectum (100%)
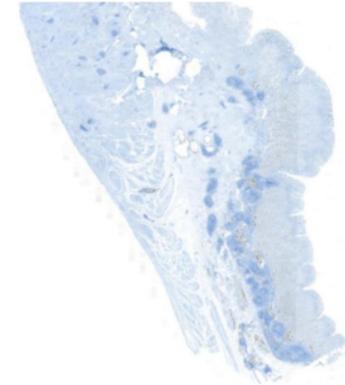

TROP2 lung (96%)
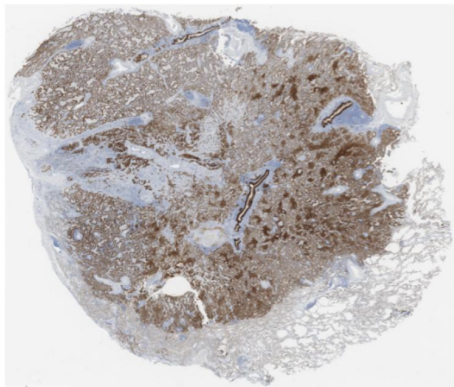

SP142 lung (100%)
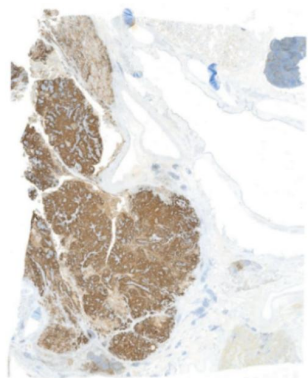

SP263 lung (43%)
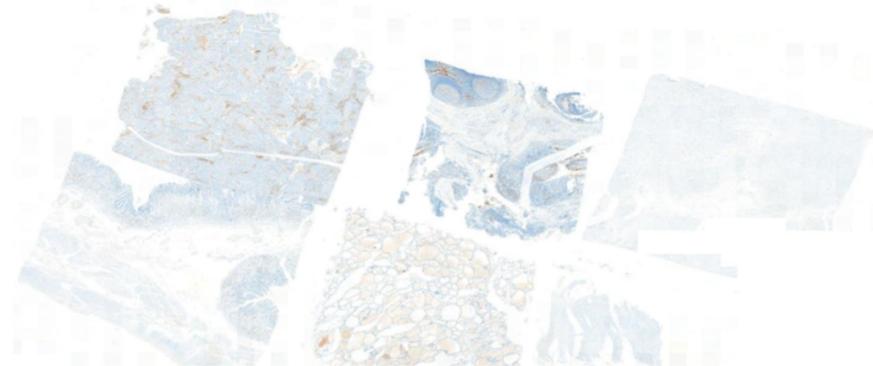